\documentclass[journal]{IEEEtran} %文献类型
\usepackage[T1]{fontenc}
\usepackage{cite} %引用
\usepackage{amsmath,amssymb,amsfonts}
\usepackage{graphicx} %插入图片
\usepackage{subcaption} % 插入子图的宏包
\usepackage{textcomp}
\usepackage{xcolor}
\usepackage{hyperref}
\usepackage{bm}
\usepackage{amsthm}
\usepackage{enumerate}
\usepackage{booktabs} %表格线条加粗
\usepackage{diagbox}
\usepackage{supertabular}
\usepackage{caption}
\usepackage{algorithmicx}
\usepackage[linesnumbered,ruled,vlined]{algorithm2e}
\usepackage{array}

\begin{document}
\title{Efficient and Trustworthy Block Propagation for Blockchain-enabled Mobile Embodied AI Networks: A Graph Resfusion Approach}

\author{Jiawen Kang, Jiana Liao, Runquan Gao, Jinbo Wen, Huawei Huang, Maomao Zhang*,\\ Changyan Yi, Tao Zhang, Dusit Niyato, \textit{Fellow, IEEE}, and Zibin Zheng, \textit{Fellow, IEEE}

% J. Kang is with the Anhui Engineering Research Center for Agricultural Product Quality Safety Digital Intelligence, Fuyang Normal University, Fuyang, 236037, China, and the School of Automation, Guangdong University of Technology, Guangzhou, 510006, China (e-mail: kavinkang@gdut.edu.cn).
\thanks{J. Kang is with the Anhui Engineering Research Center for Agricultural Product Quality Safety Digital Intelligence, Fuyang Normal University, Fuyang, 236037, China, and the School of Automation, Guangdong University of Technology, Guangzhou, 510006, China (e-mail: kavinkang@gdut.edu.cn).

J. Liao, H. Huang, and Z. Zheng are with the School of Software Engineering, Sun Yat-Sen University, Zhuhai, 519082, China (e-mails: ljn66636@163.com, huanghw28@mail.sysu.edu.cn, and zhzibin@mail.sysu.edu.cn).

R. Gao is with the School of Intelligent Systems Engineering, Sun Yat-Sen University, Shenzhen, 518107, China (e-mail: grq6012@163.com).

J. Wen and C. Yi are with the College of Computer Science and Technology, Nanjing University of Aeronautics and Astronautics, Nanjing, 210016, China (e-mails: jinbo1608@nuaa.edu.cn and changyan.yi@nuaa.edu.cn).

M. Zhang is with the Anhui Engineering Research Center for Agricultural Product Quality Safety Digital Intelligence, Fuyang Normal University, Fuyang, 236037, China (e-mail: zhangmaohi@163.com).

T. Zhang is with the School of Cyberspace Science and Technology, Beijing Jiaotong University, Beijing 100044, China (e-mail: taozh@bjtu.edu.cn).

D. Niyato is with the College of Computing and Data Science, Nanyang Technological University, Singapore (e-mail: dniyato@ntu.edu.sg).

*Corresponding author: Maomao Zhang.
}
}

\maketitle
\begin{abstract}
%摘要好像没有突出联盟链
By synergistically integrating mobile networks and embodied artificial intelligence (AI), \underline{M}obile \underline{E}mbodied \underline{A}I \underline{Net}works (MEANETs) represent an advanced paradigm that facilitates autonomous, context-aware, and interactive behaviors within dynamic environments. Nevertheless, the rapid development of MEANETs is accompanied by challenges in trustworthiness and operational efficiency. Fortunately, blockchain technology, with its decentralized and immutable characteristics, offers promising solutions for MEANETs. However, existing block propagation mechanisms suffer from challenges such as low propagation efficiency and weak security for block propagation, which results in delayed transmission of vehicular messages or vulnerability to malicious tampering, potentially causing severe traffic accidents in blockchain-enabled MEANETs. Moreover, current block propagation strategies cannot effectively adapt to real-time changes of dynamic topology in MEANETs. Therefore, in this paper, we propose a graph Resfusion model-based trustworthy block propagation optimization framework for consortium blockchain-enabled MEANETs. Specifically, we propose an innovative trust calculation mechanism based on the trust cloud model, which comprehensively accounts for randomness and fuzziness in the miner trust evaluation. Furthermore, by leveraging the strengths of graph neural networks and diffusion models, we develop a graph Resfusion model to effectively and adaptively generate the optimal block propagation trajectory. Simulation results demonstrate that the proposed model outperforms other routing mechanisms in terms of block propagation efficiency and trustworthiness. Additionally, the results highlight its strong adaptability to dynamic environments, making it particularly suitable for rapidly changing MEANETs.
\end{abstract}

\section{Introduction}
In recent years, the convergence of embodied artificial intelligence (AI) and mobile networks has led to the emergence of mobile embodied AI networks (MEANETs), which are transforming the landscape of human-computer interaction, mobile robotics, and intelligent systems\cite{zhong2025generative, wang2024embodiedscan}. Specifically, MEANETs integrate AI-driven capabilities with mobile and embedded systems, enabling mobile devices to perceive, understand, and interact with their environments in a dynamic and context-aware manner. Moreover, MEANETs focus on autonomous, intelligent systems that are capable of real-time decision-making, adaptation, and action, in unpredictable and complex real-world environments, which revolutionize traditional mobile applications. A notable application is agricultural robots, which exemplify agricultural embodied AI networks \cite{yang2024embodied, krop2024effects, kong2024embodied}. These agricultural robots leverage advanced sensing technologies, adaptive control systems, and machine learning to autonomously perform tasks such as crop monitoring, precision spraying, and harvesting. By dynamically interacting with their surroundings, agricultural robots can enhance productivity and reduce resource consumption, showcasing the transformative potential of embodied AI in addressing real-world challenges. However, considering that there are lots of distributed mobile nodes in MEANETs, which have urgent demands for decentralized resources and data sharing, it is necessary for MEANETs to integrate with blockchain to achieve secure and reliable sharing.

Blockchain technology, renowned for its decentralized, immutable, and transparent ledger system, provides a secure and efficient framework for managing data across distributed networks\cite{tripathi2023comprehensive}. When integrated with the MEANETs, consortium blockchain technology with its collaborative governance significantly enhances the security, transparency, and reliability of data transactions and network operations\cite{wang2023blockchain}. By offering a tamper-proof and verifiable platform, consortium blockchains effectively handle the vast amounts of data generated by MEANET devices (e.g., agricultural robots and vehicles), ensuring data integrity and reducing vulnerabilities. In this context, blockchain serves as the technical backbone of MEANETs, addressing critical trust issues, facilitating secure communication channels, and enabling decentralized decision-making processes. These capabilities are vital not only for enhancing operational efficiency but also for promoting the scalability, resilience, and continued growth of MEANETs.

While integrating consortium blockchain technologies into MEANETs, mobile devices in MEANETs function as miners within the consortium blockchain. However, several challenges must be addressed to develop consortium blockchain-enabled MEANETs. One of the key technical challenges is optimizing blockchain performance to meet rapid transaction demand in MEANETs. On the one hand, during block propagation, malicious miners can exploit system vulnerabilities to launch attacks, including selfish mining, Sybil attacks, and double spending, which can lead to significant financial losses, privacy breaches, and traffic paralysis, compromising the reliability of blockchains\cite{liao2024graph}. On the other hand, when block propagation time in blockchain-enabled MEANETs is excessively delayed, it can result in a higher occurrence of forks and insufficient signature collection, which may ultimately cause transaction verification failures. Moreover, the high mobility of mobile devices brings further complexity to achieving efficient block propagation. Therefore, it is critical to optimize block propagation for consortium blockchain-enabled MEANETs to ensure both performance and trustworthiness. Although some efforts have been conducted to improve the performance of block propagation\cite{liao2024graph, lee2023opportunistic, alkhalifa2024enhancing, ran2023blockchain, huang2021blockchain, wen2022optimal}, they often fail to consider the simultaneous need to optimize block propagation efficiency while improving the trustworthiness of miners in the consortium blockchain-enabled MEANETs. 

To tackle the above challenges, in this paper, we design a graph Resfusion model-based trustworthy block propagation optimization framework for consortium blockchain-enabled MEANETs. Specifically,  we first propose a trust calculation mechanism based on the trust cloud model\cite{yang2021intelligent}, comprehensively considering the ambiguity and uncertainty of miner trustworthiness, thus choosing miners with high trust scores for trustworthy block propagation. Moreover, traditional optimization methods often demonstrate suboptimal performance, while advanced deep reinforcement learning algorithms encounter challenges such as sparse rewards, leading to poor model convergence. To overcome these limitations, we propose an innovative graph Resfusion model. This model can effectively capture spatial relationships within dynamic network structures, enabling the identification of an optimal and reliable block propagation trajectory, and ultimately achieving efficient, adaptive, and trustworthy block propagation optimization. The main contributions of this paper can be summarized as follows:

\begin{itemize}
    \item Given the extensive distribution of mobile nodes in MEANETs and their demands for decentralized resources and data sharing, we integrate blockchain technologies with MEANETs and propose a trustworthy block propagation optimization framework based on the graph Resfusion model for consortium blockchain-enabled MEANETs. \textit{To the best of our knowledge, this paper is the first to tackle block optimization problems while considering miner trustworthiness in consortium blockchain-enabled MEANETs.}
    \item To achieve the trustworthiness of block propagation, we propose an innovative trust calculation mechanism based on the trust cloud model, which has an overall consideration of the randomness and fuzziness in the trust evaluation of miners. Moreover, the trust calculation mechanism considers three aspects, i.e., reputation, trustworthiness, and risk, to accurately evaluate miner trust scores, and the miners with high trust scores are selected to ensure a trustworthy block propagation trajectory. Finally, we formulate the optimization objective to minimize the total block propagation time in blockchain-enabled MEANETs with the given miner trust score constraint.
    \item Building on traditional graph diffusion methods, we propose an innovative approach that combines image-based Resfusion techniques with a gated graph neural network (GatedGNN), resulting in the development of graph Resfusion models for optimizing block propagation. Specifically, the GatedGNN integrates both edge and miner features, while the diffusion-based Resfusion technique effectively reduces the sampling space and inference time for optimal block propagation trajectories. Consequently, the proposed graph Resfusion model delivers superior block propagation performance, meeting the rapid response requirements essential for consortium blockchain-enabled MEANETs.
\end{itemize}

The remainder of the paper is structured as follows: Section \ref{related_work} provides a review of related literature. In Section \ref{system_model_section}, we present the system model, consisting of the block propagation mechanism and the designed framework of block propagation optimization. In Section \ref{cloud1}, we propose the cloud model-based trust calculation mechanism for the evaluation of miner trust scores, thereby ensuring trustworthy block propagation optimization. In Section \ref{graph}, we introduce the proposed graph Resfusion model to reveal how to obtain the optimal block propagation trajectories. Section \ref{results} provides the simulation results of the proposed schemes. Finally, the paper is concluded with Section \ref{conclusion}.
\section{Related Work}\label{related_work}
\subsection{Embodied AI and Mobile Networks}
With the rapid advancement of AI, the concept of embodied AI has emerged, highlighting the integration of intelligent systems with physical entities\cite{zhong2025generative, zhang2025embodied}. Embodied AI refers to AI systems integrated into physical entities, allowing them to perceive, interpret, and interact with their environment through sensory and motor capabilities, thereby enhancing their ability to autonomously adapt and respond to dynamic conditions\cite{wang2024embodiedscan}. Moreover, embodied AI and mobile networks are closely interconnected, as mobile networks provide the essential infrastructure for real-time communication, data exchange, and remote control. This seamless connectivity, combined with mobility and distributed intelligence, enables embodied AI systems to operate autonomously and adapt to dynamic environments\cite{zhong2025generative}. However, the security problem of mobile networks combined with embodied AI is still not well considered\cite{zhong2025generative}. Some efforts have focused on combining blockchain technologies with mobile networks to achieve trustworthy blockchain-enabled mobile networks\cite{hussain2024blockchain, liu2024hierarchical, roy2024blockchain}. For example, the authors in \cite{hussain2024blockchain} proposed a blockchain-enabled secure communication framework for autonomous vehicles, leveraging decentralized systems to enhance data integrity, trust, and access control through smart contracts and consensus mechanisms\cite{hussain2024blockchain}. The authors in \cite{liu2024hierarchical} proposed a hierarchical blockchain-enabled security-threat assessment architecture for the Internet of Vehicles, integrating edge and global chains to efficiently manage and share security-threat information, while utilizing data virtualization, a metadata association model, and ant colony optimization to reduce query delays and enhance performance\cite{liu2024hierarchical}. The authors in \cite{roy2024blockchain} proposed an innovative blockchain-based access control protocol with handover authentication for intelligent transportation systems. By integrating blockchain's decentralized storage and transparency, the protocol addressed key security challenges related to vehicle-to-vehicle and vehicle-to-infrastructure communications, while minimizing computational overhead through a lightweight handover authentication mechanism\cite{roy2024blockchain}. However, most existing studies do not consider improving the trustworthiness of miners in the block propagation process as well as ensuring the efficiency and dynamic adaptability needed to address real-time changes in mobile networks. This limitation strongly motivates the enhancement of the block propagation process for consortium blockchain-enabled MEANETs.

\subsection{Graph Diffusion Models}
In recent years, the graph diffusion model has attracted considerable attention for its ability to simulate the spread of information or influence across graph-structured data. By modeling the diffusion process, the graph diffusion model effectively captures how signals or attributes propagate through nodes and edges, making it particularly useful for applications such as social network analysis, viral marketing, and epidemic modeling. Its capacity to handle dynamic and evolving graph structures, along with its ability to model complex interactions between nodes, has made the graph diffusion model an essential tool for understanding and predicting behaviors in various networked systems\cite{jiang2024diffkg, kong2023autoregressive, huang2023conditional}. For example, the authors in \cite{jiang2024diffkg} proposed an innovative knowledge graph diffusion model for recommendations, combining generative diffusion and data augmentation techniques to improve the learning of knowledge graph representations, while incorporating collaborative knowledge graph convolution to improve user-item interaction modeling and recommendation performance \cite{jiang2024diffkg}. The authors in \cite{kong2023autoregressive} proposed an autoregressive diffusion model for graph generation that addresses the limitations of the existing one-shot diffusion models by directly operating in the discrete graph space, using a node-absorbing diffusion process and a reverse denoising network to efficiently generate graphs with improved training stability and faster sampling speed\cite{kong2023autoregressive}. The authors in \cite{huang2023conditional} introduced a novel diffusion model based on discrete graph structures for molecular graph generation, utilizing stochastic differential equations for forward diffusion, a hybrid graph noise prediction model for node-edge dependency, and efficient graph sampling with ordinary differential equation, achieving high-quality molecular graph generation with fewer steps \cite{huang2023conditional}. However, most existing work focuses on utilizing graph diffusion for graph generation or graph fusion, with little attention given to its application in optimization, particularly in the area of block propagation optimization.

\renewcommand{\arraystretch}{1.4}
\begin{table}[t]
\caption{Mathematical Notations}
% \centering
\begin{tabular}{m{0.8cm}|m{6.8cm}} %14.5
\toprule[1pt]
\hline
\multicolumn{1}{c|}{\textbf{Notation}}  & \multicolumn{1}{c}{\textbf{Definition}} \\ \hline
$Rep_i$ & Reputation value of miner $i$, reflecting the subjective evaluations of all miners, except miner $i$ \\ \hline
$TW_{r_j}$ & Trustworthiness value of block requester $r_j$, reflecting the ability of block requester $r_j$ to successfully propagate and validate a new block \\ \hline
$risk_{r_j}$ & Risk value, reflecting a latent change of reputation value and trustworthiness value\\ \hline
$TS_{r_j}$ & Final trust score of block requester $r_j$ which is based on reputation, trustworthiness, and risk value of $r_j$   \\ \hline
$V$ & Total number of miners in MEANETs \\ \hline
$\Lambda_{i-1, i}$ & Block propagation time from
miner $i - 1$ to miner $i$  \\ \hline
$\hat{G_{0}}$ & Degraded block propagation trajectory that is easily available \\ \hline
$G_0$ & Optimal block propagation trajectory \\ \hline
$T^{\prime}$ & A certain diffusion step in the diffusion process can be calculated by (\ref{get T'}) and is typically smaller than the maximum diffusion steps $T$  \\ \hline
$G_{T^{\prime}}$ & Computable block propagation trajectory that can be derived from degraded block propagation trajectory by smooth equivalence transformation techniques \\ \hline
$res\epsilon$ & Residual noise between the degraded input block propagation trajectory and the optimal propagation trajectory \\ \hline
$\boldsymbol{v}_i^r$ & Embedding of miner node $i$ in the $r$-th GatedGNN layer \\ \hline
$\boldsymbol{e}_{ij}^r$ & Embedding of the edge between miner $i$ and miner $j$ in the $r$-th GatedGNN layer \\ \hline
\bottomrule[1pt]
\end{tabular}
\end{table}

\subsection{Block Propagation Optimization}
The enhancement of block propagation performance constitutes a critical avenue of inquiry in the pursuit of advancing blockchain performance. Extensive research efforts have been devoted to this domain, which can be systematically classified into three overarching methodological paradigms\cite{lee2023opportunistic, alkhalifa2024enhancing, ran2023blockchain, huang2021blockchain, wen2022optimal}: 1) Optimizing the block verification: the authors in \cite{lee2023opportunistic} introduced an opportunistic block validation mechanism for IoT blockchain networks, optimizing block validation through a lightweight blockchain framework that evaluates node reputation, validation degree, and network stability using reinforcement learning while ensuring transaction data integrity and practicality for low-resource IoT devices\cite{lee2023opportunistic}. 2) Optimizing the blockchain network topology: the authors in \cite{alkhalifa2024enhancing} proposed a Bayesian DAG blockchain optimized for 5G-enabled vehicular networks, leveraging edge-assisted roadside units and grid-based topology optimization to enhance scalability and reduce computational complexity in decentralized authentication\cite{alkhalifa2024enhancing}. The authors in \cite{ran2023blockchain} proposed a blockchain topology optimization method based on node clustering, aimed at increasing transaction throughput and improving scalability by redesigning the consensus and message propagation mechanisms to mitigate delays caused by node verification and consensus processes in large-scale blockchain systems\cite{ran2023blockchain}. 3) Optimizing the block propagation behavior: the authors in \cite{huang2021blockchain} proposed a blockchain network propagation mechanism based on the P4P architecture, leveraging network topology and link status to optimize node connections, prioritizing high-bandwidth paths, and enhancing propagation speed, which can reduce fork probability and improve the quality of services in blockchain networks\cite{huang2021blockchain}. The authors in \cite{wen2022optimal} addressed block propagation challenges in 6G-enabled blockchain networks by introducing an epidemic-inspired block propagation model and an incentive mechanism based on evolutionary game theory to minimize propagation delay, demonstrating improved efficiency and stronger incentives compared with traditional routing algorithms\cite{wen2022optimal}. However, limited research has addressed the integration of miner trustworthiness with block propagation efficiency to optimize the block propagation process. Moreover, most existing studies cannot cope with the influence of dynamic change of miner networks on block propagation. In MEANETs, blockchain technology holds substantial potential to enhance the timely and trustworthy interaction between mobile embodied AI agents. Thus, it remains a challenging endeavor to achieve trustworthy and efficient block propagation optimization.

Motivated by the aforementioned research challenges, we propose a novel graph Resfusion model-based trustworthy block propagation optimization framework to enhance the performance of consortium blockchain-enabled MEANETs, in which we propose the trust calculation mechanism to ensure the trustworthiness of block propagation and the graph Resfusion model to elevate the efficiency of block propagation.
\section{System Model}\label{system_model_section} 
\begin{figure}
    \centering
    \includegraphics[width=1\linewidth]{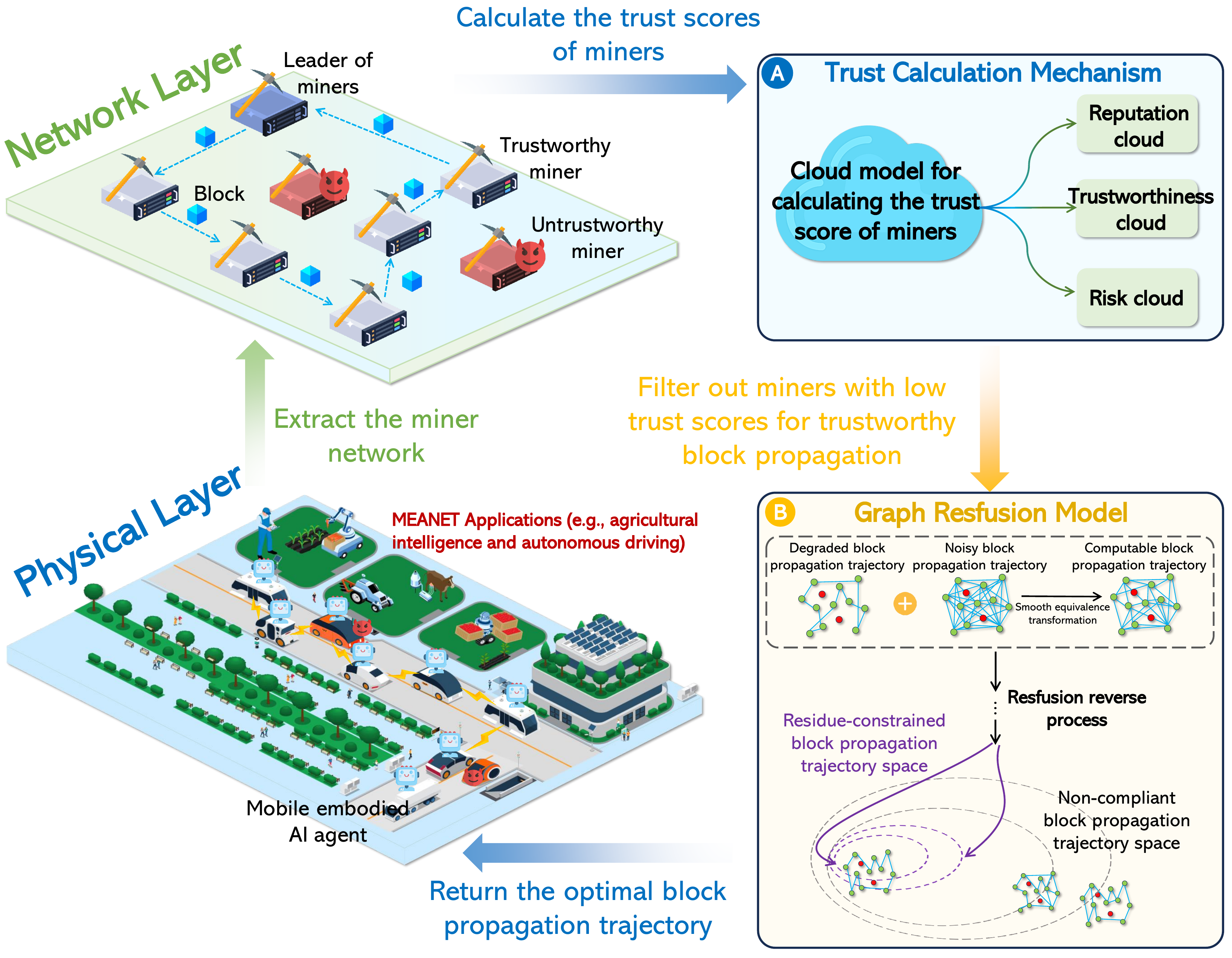}
    \caption{A graph Resfusion model-based trustworthy block propagation optimization framework for consortium blockchain-enabled MEANETs.}
    \label{system_model}
\end{figure}

\subsection{Block Propagation Mechanism}\label{mechanism}
In this section, we introduce a block propagation mechanism in consortium blockchains. Firstly, a block is generated based on consensus mechanisms (i.e., practical byzantine fault tolerance (PBFT)). During this phase, untrusted miner nodes are identified and excluded from the propagation process, as detailed in Section \ref{cloud1}. Once untrusted miners are filtered out, the block is disseminated among the remaining trusted miner nodes for verification. Specifically, the remaining propagation process can be modeled as the traveling salesman problem (TSP) in terms of miner node selection alone\cite{li2017consortium}, where the block keeper (also termed the miner leader) sequentially forwards the block to each participating miner node. After each node has validated the block and submitted its vote, the block is returned to the original miner, which consolidates and reports the verification results. Ultimately, the verification results determine whether the block is eligible to be added to the blockchain.

\subsection{Framework Design}
As illustrated in Fig. \ref{system_model}, we design the efficient and trustworthy block propagation framework for blockchain-enabled MEANETs with a two-layer structure, i.e., the physical layer and the network layer. The physical layer consists of diverse applications of MEANETs, such as mobile embodied AI agents, including vehicles, agricultural robots, and mobile phones, which are powered by lightweight large language models. MEANETs typically involve significant data transmission and sharing, raising concerns about the security and privacy of the transmitted data. When dealing with sensitive or critical information, malicious attacks can lead to severe consequences, such as information leakage, traffic accidents, or machine interaction failures. To mitigate these risks, the integration of blockchain technology is essential. The network layer, abstracted from the physical layer, presents the miner network in consortium blockchains-enabled MEANETs. The block propagation process is according to the mechanism detailed in Section \ref{mechanism}. Here, the mobile embodied AI agents serve as the miners in consortium blockchains. To ensure reliability in the block propagation, we introduce a cloud model-based trust calculation mechanism to evaluate the trustworthiness of miners\cite{huang2023trust}. This mechanism accounts for the inherent randomness and fuzziness in trust evaluation and integrates three components: the reputation cloud, the trustworthiness cloud, and the risk cloud. Miners with high trust scores are selected for participation in the trustworthy block propagation process. Subsequently, we propose the graph Resfusion model to determine the optimal block propagation trajectory\cite{zhenning2023resfusion}. We can obtain a computable block propagation trajectory as the starting point of the Resfusion reverse process from an easily available degraded block propagation trajectory by applying smooth equivalence transformation techniques. A key advantage of the Resfusion model lies in its ability to generate a residue-constrained block propagation trajectory space. This ability can avoid non-compliant trajectories, which results in reduced inference time and enhanced inference accuracy during the denoising process.

\section{Cloud Model-based Trust Calculation Mechanism for Trustworthy Block Propagation Optimization}\label{cloud1}
The cloud model, introduced by De-Yi Li \cite{li2017artificial}, provides a unified framework to represent both randomness and fuzziness in trust relationships, leveraging random mathematics and fuzzy set theory. It has been successfully applied in various trust-related fields, including subjective trust modeling and assessment. Moreover, the trust scores of miners in blockchain-enabled MEANETs have the characteristics of randomness and fuzziness\cite{ji2024dependence}. Therefore, to achieve a convincing trust score calculation for miners in consortium blockchain-enabled MEANETs, we utilize the trust cloud model-based trust calculation mechanism as an effective tool to measure the trustworthiness of miners.

We consider a set ${\mathcal{V}}=\{1,\ldots, i,\ldots,j,\ldots, V\}$ of $V$ miners with random mobility in mobile miner networks. Moreover, we consider that miner $i$ is a block provider, denoted as $p_i$, and the action of $p_i$ is to forward the holding block, denoted as $a_i$. The miner $j$ is a block requester, denoted as $r_j$, and the objective of $r_j$ is to request a new block from $p_i$, denoted as $o_j$. Therefore, the block propagation process is denoted as <$r_j, o_j, p_i, a_i$>. Specifically, the trustworthy block propagation process can be modeled as a trust-decision-making process (TDMP), termed as $\mathrm{TDMP}(r_j(o_j), p_i(a_i))$\cite{wang2010cloud}. 

In $\mathrm{TDMP}(r_j(o_j), p_i(a_i))$, we propose the trust cloud model-based trust calculation mechanism to calculate the trust score of miners in consortium blockchain-enabled MEANETs by considering three aspects, i.e., reputation, trustworthiness, and risk.

\subsection{Preliminary Work of Cloud and Cloud Drops}
Let $\mathcal{U}$ denote the universal set characterized by precise values, and let $C$ denote a qualitative concept associated with $\mathcal{U}$. Suppose a number $x \in U$  randomly embodies the concept \( C \), with a certainty degree \( \mu(x) \in [0,1] \) that reflects the likelihood of \( x \) conforming to \( C \). The certainty degree \( \mu(x) \) is a random variable with a tendency toward stability\cite{wang2010cloud}:
\begin{equation}\label{cloud}
\mu : \mathcal{U} \rightarrow [0,1], \quad \forall x \in \mathcal{U},
\end{equation}
\begin{equation}
\quad x \rightarrow \mu(x).    
\end{equation}
Based on (\ref{cloud}), the distribution of \( x \) over $\mathcal{U}$ is referred to as a cloud, and each instance of \( x \) is termed a cloud drop. Moreover, the certainty degree of \( x \) corresponds to the membership degree in fuzzy set theory but follows a probability distribution, rendering it a random variable, as opposed to the fixed values in conventional fuzzy membership functions.

\begin{figure}
    \centering
    \includegraphics[width=0.9\linewidth]{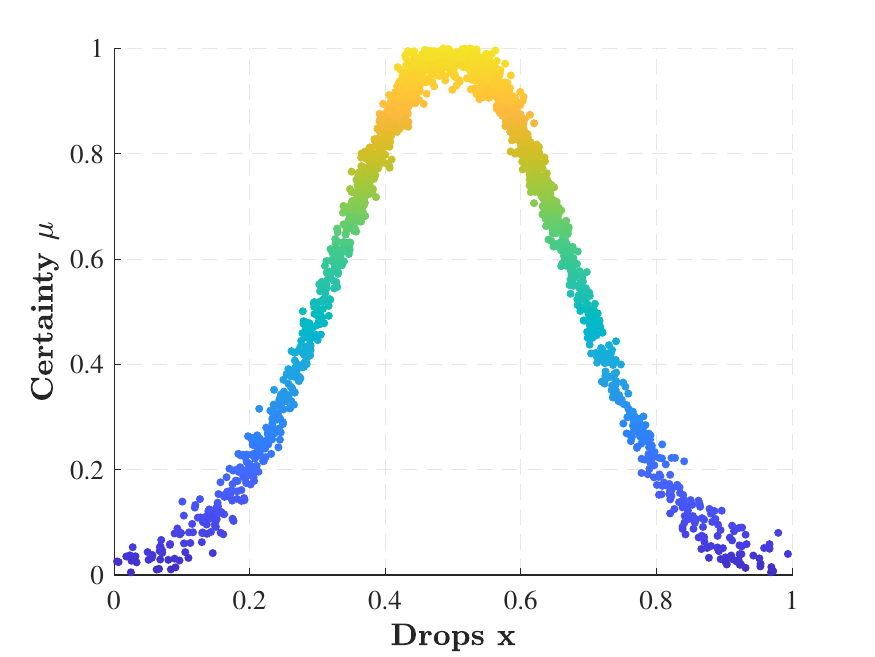}
    \caption{A one-dimensional normal cloud model.}
    \label{22}
\end{figure}

The cloud model characterizes a subjective concept through three numerical attributes: the expected value \( Ex \), the entropy \( En \), and the hyper-entropy \( He \). Collectively, $Ex$, $En$, and $He$ are termed the numerical characteristics of a cloud. \( Ex \) represents the average position of cloud drops within $\mathcal{U}$, serving as a central indicator of the qualitative concept. \( En \) quantifies the uncertainty associated with the concept, capturing both randomness and fuzziness. Meanwhile, \( He \) measures the uncertainty inherent in \( En \), reflecting its variability due to the combined influence of randomness and fuzziness. Specifically, $Ex$, $En$, and $He$ are given by\cite{ji2024dependence} 
\begin{equation}\label{ex}
Ex=\frac{\sum_{i=1}^{n}{\boldsymbol{x}_i}}{n},
\end{equation}
\begin{equation}\label{en}
E{n}=\sqrt{\frac{\pi}{2}}\times\frac{1}{n}\sum_{i=1}^{n}|\boldsymbol{x}_{i}-\bar{x}|,\end{equation}
\begin{equation}
B=\frac{\sum_{i=1}^{n}\left(\boldsymbol{x}_{i}-\bar{x}\right)^{2}}{n},
\end{equation}
\begin{equation}\label{he}
He=\sqrt{{B}^{2}-E{n}^{2}},
\end{equation}
where $\boldsymbol{x}_{i}$ denotes the $i$ cloud drop, $n$ denotes the number of the cloud drops, $\bar{x}$ denotes the average of the total number of $\boldsymbol{x}_i$, and $B$ denotes the variance of cloud drops. Figure \ref{22} demonstrates a one-dimensional normal cloud model, where $Ex=0.5$, $En=0.15$, and $He=0.03$\cite{wang2010cloud}. These values represent a system with an expected value of $0.5$, exhibiting low systematic and minimal entropy uncertainty.

\subsection{Reputation Calculation for Miners}
In cloud model-based $\mathrm{TDMP}(r_j(o_j), p_i(a_i))$, the reputation reflects \textit{the subjective evaluations of all miners after interacting with $r_j$}, primarily based on positive or negative experiences and subjective perceptions within the miner network. Furthermore, reputation serves as a recommended opinion, denoted as $Rep_i\in [0,1]$.

Specifically, we define the reputation value $Rep_i$ of miner $i$ that acts as block provider $p_i$ forwarding the block to block requester $r_j$ as follows\cite{huang2023trust}:
\begin{equation}Rep_i=\frac{o_i}{o_i + n_i}+h_i,\: i\in\{1, \ldots, V\}, i \neq r_j,\label{rep}\end{equation}
where $o_i$ and $n_i$ denote the numbers of positive and negative miner interactions between block provider $p_i$ and block requester $r_j$, respectively. $h_i\in[0, \eta]$ is a subjective evaluation value from block provider $p_i$ to block requester $r_j$, which is subject to the block waiting time that refers to the duration a newly generated block remains in a node's queue before being propagated. If the block waiting time is below a predefined threshold $t_{wait}$, $h_i$ can be assigned a value close to $\eta$, where $\eta$ is set to $0.2$ indicating that $h_i$ can contribute up to one-fifth of the total reputation value, thereby incentivizing favorable transmission bandwidth. Conversely, if the block waiting time exceeds this threshold, $h_i$ can be set to a value close to $0$, indicating minimal or no reward allocation.

Furthermore, based on (\ref{cloud}), the cloud model of the reputation of miners can be given by\cite{wang2010cloud}
\begin{equation}
  \mu : Rep_i \rightarrow [0,1], \quad \forall x \in Rep_i, i \in \{1, \ldots, V\}, i \neq r_j, 
\end{equation}
\begin{equation}
  x \rightarrow \mu(x),
\end{equation}

where the distribution of $x$ in $Rep_i$ is termed the reputation cloud, denoted as $RepC(x)$, and each $x$, calculated by (\ref{rep}), is referred to as a reputation cloud drop.
\subsection{Trustworthiness Calculation for Miners}
In trust cloud model-based $\mathrm{TDMP}(r_j(o_j), p_i(a_i))$, 
the trustworthiness, denoted as $TW_{r_j}$, \textit{depends on the ability of the miner block requester $r_j$ to successfully validate and forward the new block.} $TW_{r_j}$ is calculated as the arithmetic average of the values of all relevant capability indicators. Specifically, we model the trustworthiness based on three factors, called the average miner block propagation time, the miner communication link quality, and the ratio of the number of active external interactions of the block requester $r_j$ to the total number of external interactions of $r_j$. Furthermore, the abilities can be categorized as either positive or negative. A positive ability indicator is a metric where a higher value is advantageous, such as the average miner block propagation time and the miner communication link quality. In contrast, a negative indicator is a metric where a lower value is preferable, such as the ratio of the number of active external interactions of the block requester $r_j$ to the total number of external interactions of $r_j$.

Specifically, we denote the $y$-th ability of $r_j$ as $Cap_{r_j}^y$, where $y\in[1, \ldots, Y]$ and $Y=3$. For notational simplicity, $Cap_{r_j}^y$ can be expressed as $Cap_{r_j}$, which can be calculated as
\begin{itemize}
\item \textbf{\emph{For negative ability indicators:}}
    \begin{equation}\label{ne}
    Cap_{r_j}=\frac{Cap_{min}-Cap_{r_j}}{Cap_{max}-Cap_{min}}, r_j \in \{1, \ldots, V\}.\end{equation}
\item \textbf{\emph{For positive ability indicators:}}
    \begin{equation}\label{po}
    Cap_{r_j}=\frac{Cap_{r_j}-Cap_{min}}{Cap_{max}-Cap_{min}}, r_j \in \{1, \ldots, V\}.\end{equation}
\end{itemize}
Here $Cap_{min}$ and $Cap_{max}$ denote the minimum and maximum of the $y$-th ability of miners $r_j$, respectively. The objective of (\ref{ne}) and (\ref{po}) is to scale the $Cap_{r_j}$ into $[0,1]$. 
Therefore, the total trustworthiness value can be given by
\begin{equation}\label{tw1}
    TW_{r_j} = \sum_{y=1}^{Y}\boldsymbol{w}_yCap_{r_j}^y, r_j \in \{1, \ldots, V\},
\end{equation}
\begin{equation}\label{tw2}
    \sum_{y=1}^{Y} \boldsymbol{w}_y = 1,
\end{equation}
where $w_y$ denotes the weight of each ability of miners $r_j$.

Furthermore, based on (\ref{cloud}), the cloud model of miner trustworthiness can be given by
\begin{equation}
  \mu : TW_{r_j} \rightarrow [0,1], \quad \forall x \in TW_{r_j}, r_j \in \{1, \ldots, V\},  \quad 
\end{equation}
\begin{equation}
  x \rightarrow \mu(x),
\end{equation}
where the distribution of $x$ in $TW_{r_j}$ is termed the trustworthiness cloud, denoted as $\mathrm{TWC}(x)$, and each $x$, calculated by (\ref{tw1}) and (\ref{tw2}), is referred to as a trustworthiness cloud drop.
\subsection{Risk Calculation for Miners}
In the cloud model-based $\mathrm{TDMP}(r_j(o_j), p_i(a_i))$, the risk of miner $r_j$, $risk_{r_j}$, \textit{is defined as a latent change in reputation and trustworthiness}. To calculate $risk_{r_j}$, we first calculate $R_{dif}$ that determines the difference in reputation or trustworthiness between adjective time slot, which is given by\cite{huang2023trust}
\begin{equation}\label{Rdif}
    R_{dif}=x_{i'}-x_i,
\end{equation}
where $x_{i'}$ denotes the reputation or trustworthiness value in the next time slot, and the value of $R_{dif}$ could be positive or negative. Moreover, we define $R_l$ and $R_u$ as the lower and upper bounds of $R_{dif}$, respectively. Since the range of both reputation and trustworthiness is $[0,1]$, $R_l$ and $R_u$ are set to $-1$ and $1$, respectively. Therefore, based on (\ref{Rdif}), the final calculation of $risk_{r_j}$ is given by
\begin{equation}\label{ri1}
risk_{r_j}=\frac{R_{dif}-R_l}{R_u-R_l},
\end{equation}
where $risk_{r_j}$ can be scaled into $[0,1]$.

Furthermore, based on (\ref{cloud}), the cloud model of miner risk can be given by
\begin{equation}
  \mu : risk_{r_j} \rightarrow [0,1], \quad \forall x \in risk_{r_j}, {r_j} \in \{1, \ldots, V\},  \quad 
\end{equation}
\begin{equation}
  x \rightarrow \mu(x),
\end{equation}
where the distribution of $x$ in $risk_{r_j}$ is termed the risk cloud, denoted as $\mathrm{RiskC}(x)$, and each $x$, calculated by 
(\ref{ri1}), is referred to as a risk cloud drop.
\subsection{Final Trust Score Calculation for Miners}
Reputation and trustworthiness meet the subjective uncertainty and fuzziness of cloud models. According to \cite{wang2010cloud}, the computable trust scores of reputation or trustworthiness of miner $r_j$ can be given by
\begin{equation}\label{s}
S=Ex\times e^{-He} + Ex,
\end{equation}
where the expected value $Ex$, the entropy $En$, and the hyper-entropy $He$ can be calculated by (\ref{ex}), (\ref{en}), and (\ref{he}), respectively. (\ref{s}) is designed for both reputation and trustworthiness based on the established cloud model. For clarity, we denote $S$ for reputation as $S_{rep}$ and the $S$ for trustworthiness as $S_{tw}$.

Based on the value of $S_{rep}$ and $S_{tw}$ calculated by (\ref{s}) and considering the impact of reputation, trustworthiness, and risk comprehensively, we can obtain the final trust score of block requester $TS_{r_j}$ as \cite{wang2010cloud}
% where $Ex$ denotes the expectation in the cloud model, reflecting the concept's central point or general value, and $En$ denotes the fuzziness and randomness of the concept. The larger the $En$ value, the vaguer the boundary of the concept, that is, the wider the distribution of the data in the concept. The smaller the $En$ value, the clearer the definition of the representation concept and the more concentrated the data. $He$ super-entropy is the entropy of entropy, reflecting the uncertainty of entropy, describing the degree of entropy fluctuation within the concept.
\begin{equation}
TS_{r_j}=\frac{S_{rep}+S_{tw}-e^{\mathrm{RiskC}(S_{rep})}-e^{\mathrm{RiskC}(S_{tw})}}{2\times e},
\end{equation}
where $r_j\in\{1,\ldots,V\}$ and $e$ denotes the natural number $e$. The $\mathrm{RiskC}(S_{rep})$ and the $\mathrm{RiskC}(S_{tw})$ denote the $risk_{r_j}$ calculation for reputation and trustworthiness of block requester $r_j$, respectively. Finally, $TS_{r_j}$ will be processed through a normalization step to scale the range in $[0,1]$. After understanding the calculation of the trust scores of the miners, we will introduce the proposed graph Resfusion model-based architecture for block propagation optimization in the following section.

\begin{figure*}
    \centering
    \includegraphics[width=1\linewidth]{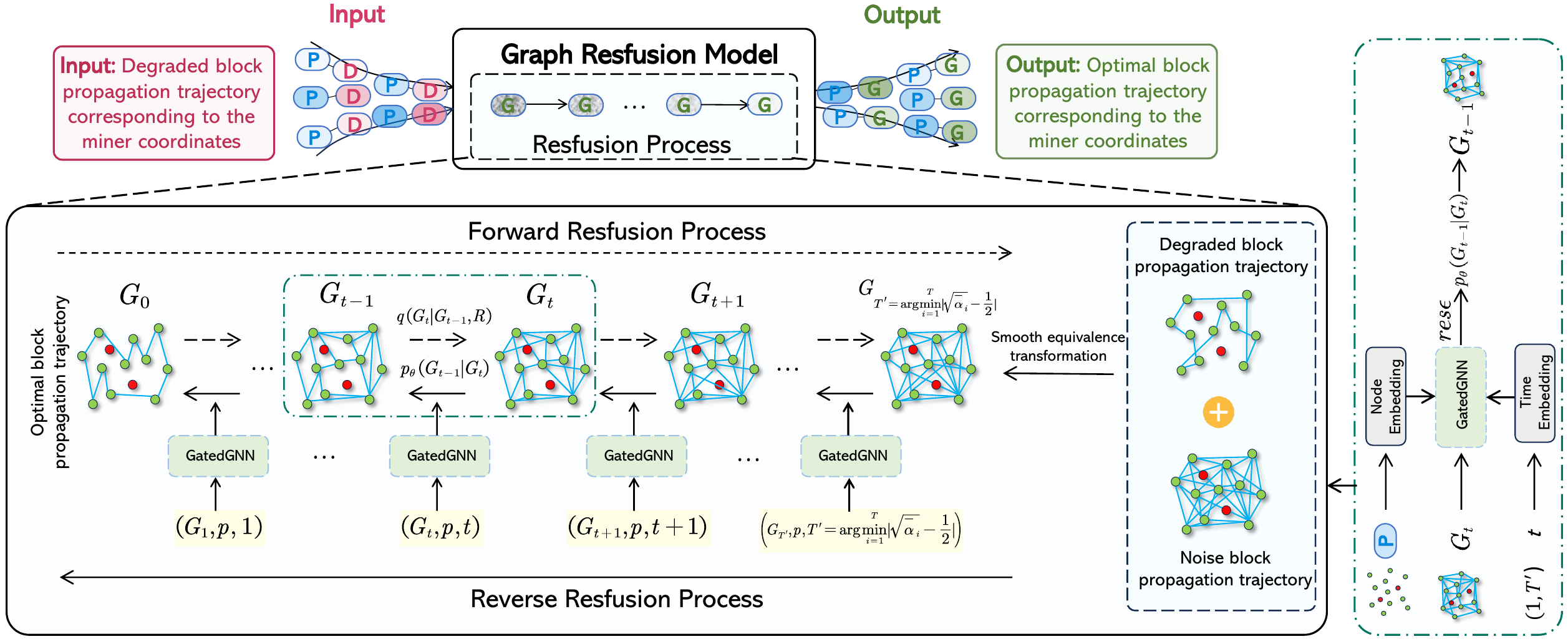}
    \caption{The comprehensive architecture of the graph Resfusion model. This model generates the optimal block propagation trajectory by inputting the easily available degraded block propagation trajectory under the corresponding miner coordinates, thereby empowering the consortium blockchain-enabled MEANETs shown in Fig. \ref{system_model}.}
    \label{resfusion}
\end{figure*}

\section{Graph Resfusion Model-based Solution for Block Propagation Optimization}\label{graph}
In this section, we address the challenge of optimizing trustworthy block propagation in consortium blockchains by identifying the optimal block propagation trajectory utilizing the heatmap derived from the graph Resfusion model, and the comprehensive architecture of the proposed graph Resfusion model is depicted in Fig. \ref{resfusion}.

\subsection{Optimization Goal}
Building upon the PBFT consensus mechanism adopted in consortium blockchains, the optimization objective is to minimize the total block propagation time with the given trust score constraint, which is calculated by the Shannon formula as follows:
\begin{equation} 
\begin{split}
    &\min\:\sum_{i=2}^{K}\Bigg( \Lambda_{i-1, i}=\frac{S_{block}}{W\log_2\Big(1+\frac{a \rho \Gamma_{i-1,i}^{-\varepsilon}}{N_0W}\Big)}\Bigg)+\Lambda_{K,0}\\
    &\:\:\mathrm{s.t.}\:TS_i>\lambda, i\in \{1,\ldots,K\},
\label{fu}
\end{split}
\end{equation}
where $\Lambda_{i-1, i}$ denotes the time taken for block propagation from miner $i-1$ to miner $i$, and $\Lambda_{K,0}$ signifies that the block must be propagated to the first block provider. $i\in\{1,\ldots,K\}$ represents a subset of $\mathcal{V}$, representing the $(V-K)$ mining nodes removed from $\mathcal{V}$. To ensure secure block propagation, each miner's trust score in set $\mathcal{S}$ must exceed a threshold $\lambda \in(0,1)$.  $S_{block}$, $W$, and $\rho$ refer to block size, channel bandwidth between adjacent miners, and transmit power, respectively.  Moreover, ${\Gamma_{i-1, i}}$, $N_0$, $\varepsilon$, and $a$ refer to block propagation distance between miner $i-1$ and miner $i$, noise power density, path loss exponent, and unit channel power gain, respectively. 

The following section presents the proposed graph Resfusion model in detail, including the forward Resfusion process, reverse Resfusion process, and GatedGNN for Resfusion.

\subsection{Forward Resfusion Process}
Unlike previous forward processes \cite{sun2023difusco}, \cite{ho2020denoising} in diffusion models, the Resfusion forward process introduces the concept of residuals. First, we define the degraded block propagation trajectory and the optimal block propagation trajectory as $\hat{G_{0}}$ and $G_{0}$, respectively. The degraded block propagation trajectory refers to a trajectory that satisfies problem constraints but has not been specially optimized. In practice, these trajectories are easily available and can be generated by connecting each miner in sequence or using other simple algorithms. This approach of deriving high-quality solutions from low-quality ones has proven effective and is widely applied\cite{zhang2024deep}. Accordingly, the residual is defined as follows:
\begin{equation}
R=\hat{G_{0}}-G_{0}.\label{define res}
\end{equation}

Subsequently, the Resfusion forward process with residuals can be succinctly denoted as $(G_{0}, G_{1},\ldots, G_{T})$, where $T$ denotes the maximum diffusion step. The forward process can be formalized as follows:
\begin{equation}
q(G_{1:T}|G_0,R)=\prod\limits_{t=1}^Tq(G_t|G_{t-1},R),\label{forward 0 to 1:T}
\end{equation}
\begin{equation}
q(G_t|G_{t-1},R)=\mathcal{N}(G_t;\sqrt{\alpha_t}G_{t-1}+(1-\sqrt{\alpha_t})R,(1-\alpha_t)I).\label{forward t-1 to t}
\end{equation}

Here, $\alpha_t=1-\beta_t$, where $\{\beta_{1}<\beta_{2}<\cdots<\beta_{T},\beta_{t}\in(0,1)\}$ represents a fixed schedule that controls the level of noise and residual added at each diffusion step. (\ref{forward 0 to 1:T}) and (\ref{forward t-1 to t}) demonstrate that the Resfusion forward process gradually incorporates residuals and noise into the optimal block propagation trajectory $G_{0}$, leading the final block propagation trajectory toward disorder.

(\ref{forward t-1 to t}) represents the Resfusion forward process from $t-1$ diffusion steps to $t$ steps, which can be rewritten as
\begin{equation}
G_t=\sqrt{\alpha_t}G_{t-1}+(1-\sqrt{\alpha_t})R+\sqrt{1-\alpha_t}\epsilon,\label{forward 0 to t}
\end{equation}
where $\epsilon\sim N(0, I)$ denotes Gaussian noise. 

\begin{algorithm}[t!]
    \small
    \caption{Training Algorithm for Forward Resfusion Process}\label{Traning algorithm}
    
    \KwIn{Total diffusion step $T$; Training epoch $E$; Degraded block propagation trajectory and optimal block propagation trajectory $(\hat{G_{0}},G_{0}).$ }\label{input1}
    \KwOut{Well trained graph neural network $res\epsilon_\theta(\cdot,\cdot,\cdot).$} \label{output1}

    \ Get diffusion step $T^{\prime}=\arg\min_{i=1}^T|\sqrt{\overline{\alpha}_i}-\frac{1}{2}|$\;
    \For{$j = 1,2,\dots,E$ \do}{
        \ Sample $\epsilon\sim\mathcal{N}(0,I)$\;
        \ Sample $t\sim \textrm{Uniform}(\{1,\ldots,T^{\prime}\})$\;
        \ $R=\hat{G_{0}}-G_{0}$\;
        \ $G_{t}=\sqrt{\overline{\alpha}_{t}}G_{0}+(1-\sqrt{\overline{\alpha}_{t}})R+\sqrt{1-\overline{\alpha}_{t}}\epsilon$\;
        \ $res\epsilon=\epsilon+\frac{(1-\sqrt{\alpha_t})\sqrt{1-\overline{\alpha}_t}}{\beta_t}R$\;
        \ Compute the loss function $\mathcal{L}=||res\epsilon-res\epsilon_\theta(G_t,p,t)||^{2}$ and take the gradient descent step. 
    }
    
\end{algorithm}

\begin{algorithm}[t]
    \small
        \caption{Sampling Algorithm for Reverse Resfusion Process}\label{Sampling algorithm}
    
    \KwIn{Total diffusion step $T$; Degraded block propagation trajectory  $\hat{G_{0}}$; Well trained graph neural network $res\epsilon_\theta(\cdot,\cdot,\cdot).$}\label{input}
    \KwOut{Final block propagation trajectory $G_{0}.$} \label{output}

    \ Get diffusion step $T^{\prime}=\arg\min_{i=1}^T|\sqrt{\overline{\alpha}_i}-\frac{1}{2}|$\;
    \ Sample $\epsilon\sim\mathcal{N}(0,I)$\;
    \ Calculate block propagation trajectory at step $T^{\prime}$ using: $G_{T^{\prime}}=\sqrt{\overline{\alpha}_{T^{\prime}}}\hat{G}_{0}+\sqrt{1-\overline{\alpha}_{T^{\prime}}}\epsilon.$\;
    \For{$t = T^{\prime},T^{\prime}-1,\ldots,2 $ \do}{
        \ Sample $z\sim\mathcal{N}(0,I)$\;
        \ Calculate the blockchain propagation trajectory at step $t-1$ using the following formula: 
        $\begin{aligned}
        G_{t-1} & =\Bigg(\frac{1}{\sqrt{\alpha_{t}}}(G_{t}-\frac{\beta_{t}}{\sqrt{1-\overline{\alpha}_{t}}}res\epsilon_{\theta}(G_{t},p,t))+\\
        \sqrt{\widetilde{\beta}_{t}}z\Bigg);
        \end{aligned}$\
    }
    \ Calculate the final block propagation trajectory using: $G_{0}=\Bigg(0.5(\frac{1}{\sqrt{\alpha_{1}}}(G_{1}-\frac{\beta_{1}}{\sqrt{1-\overline{\alpha}_{1}}}res\epsilon_{\theta}(G_{1},p,1))+1)\Bigg).$\
\end{algorithm}

By applying the reparameterization trick and denoting $\overline{\alpha}_{t}=\prod_{i=1}^t\alpha_t$, the block propagation trajectory $G_t$ at any diffusion step $t$ with added noise and residuals can be calculated in a closed form from the optimal block propagation trajectory $G_0$:
\begin{equation}
G_{t}=\sqrt{\overline{\alpha}_{t}}G_{0}+(1-\sqrt{\overline{\alpha}_{t}})R+\sqrt{1-\overline{\alpha}_{t}}\epsilon.\label{forward x0 to xt}
\end{equation}

\subsection{Reverse Resfusion Process}
The Resfusion reverse process $(G_T,G_{T-1},\ldots,G_0)$ is defined as follows:
\begin{equation}
p_\theta(G_{0:T-1}|G_{T})=\prod_{t=1}^{T}p_\theta(G_{t-1}|G_t).
\end{equation}

However, with an introduction of residuals, the block propagation trajectory $G_T$ at diffusion step $T$ may no longer be the Gaussian noise in normal denoising diffusion models. Therefore, determining the starting point of the reverse process in Resfusion becomes a key issue to address.

Fortunately, based on (\ref{define res}) and (\ref{forward x0 to xt}), the closed form can be further reformulated as
\begin{equation}
G_{t}=(2\sqrt{\overline{\alpha}_{t}}-1)G_{0}+(1-\sqrt{\overline{\alpha}_{t}})\hat{G}_{0}+\sqrt{1-\overline{\alpha}_{t}}\epsilon.\label{forward x0 to xt w/o R}
\end{equation}

%想一下怎么解释T^'的物理含义/ T^'可以被看作是反向过程的新起点，它通常比原来的反向过程的起点T小得多。
It can be observed through (\ref{forward x0 to xt w/o R}) that the weight coefficient $(2\sqrt{\overline{\alpha}_{t}}-1)$ of $G_0$ can be very close to 0. Thus, according to \cite{zhenning2023resfusion}, we can apply a smoothing equivalent transformation technique to obtain a computable block propagation trajectory $G_{T^{\prime}}$ at diffusion step $T^{\prime}$ with small bias. $T^{\prime}$ can be regarded as the new starting point of the reverse process, which is typically much smaller than the maximum diffusion step $T$. The definition of $T^{\prime}$ and the calculation of $G_{T^{\prime}}$ are as follows:
\begin{equation}
T^{\prime}=\arg\min_{i=1}^T|\sqrt{\overline{\alpha}_i}-\frac{1}{2}|,\label{get T'}
\end{equation}
\begin{equation}
G_{T^{\prime}}\approx\sqrt{\overline{\alpha}_{T^{\prime}}}\hat{G}_{0}+\sqrt{1-\overline{\alpha}_{T^{\prime}}}\epsilon.\label{get G_T' as start point}
\end{equation}

The derivation of $T^{\prime}$ indicates that we only need to train the forward process and execute the reverse process from $0$ to $T^{\prime}$. Thus, given the computable block propagation trajectory $G_{T^{\prime}}$ at diffusion step $T^{\prime}$, the Resfusion reverse process can be redefined as% 这里的均值和方差考虑不带西塔，后面说明了res的时候再加上西塔
\begin{equation}
p_\theta(G_{0:T^{\prime}-1}|G_{T^{\prime}})=\prod_{t=1}^{T^{\prime}}p_\theta(G_{t-1}|G_t),
\end{equation}
\begin{equation}
p_\theta(G_{t-1}|G_t)=\mathcal{N}(G_{t-1};\mu_\theta(G_t,p, t),\Sigma_\theta(G_t,p,t)).\label{mean and var}
\end{equation}

Similar to \cite{zhenning2023resfusion}, we fix the variance at each step of the reverse process as $
\Sigma_\theta(G_t,p,t)=\widetilde{\beta}_{t}=\frac{1-\overline{\alpha}_{t-1}}{1-\overline{\alpha}_{t}}\beta_{t}$, and the mean $\mu_{\theta}(G_{t},p,t)$ is computed as follows:
\begin{equation}
\mu_{\theta}(G_{t},p,t)=\Bigg(\frac{1}{\sqrt{\alpha_{t}}}(G_{t}-\frac{\beta_{t}}{\sqrt{1-\overline{\alpha}_{t}}}res\epsilon_{\theta})\Bigg).\label{mean in reverse process}
\end{equation}

Here, $res\epsilon_{\theta}$ represents the residual noise in the reverse Resfusion process. Since $res\epsilon_{\theta}$ cannot be directly obtained during the reverse process, we use a graph neural network to estimate $res\epsilon_{\theta}$, thereby deriving the mean in (\ref{mean in reverse process}). The details of the graph neural network are shown in Section \ref{GNN}. Meanwhile, to train the graph neural network, we define a directly computable residual noise $res\epsilon$ in the forward process as follows\cite{zhenning2023resfusion}:
\begin{equation}
res\epsilon=\epsilon+\frac{(1-\sqrt{\alpha_t})\sqrt{1-\overline{\alpha}_t}}{\beta_t}R.
\end{equation}

\begin{figure}
    \centering
    \includegraphics[width=0.9\linewidth]{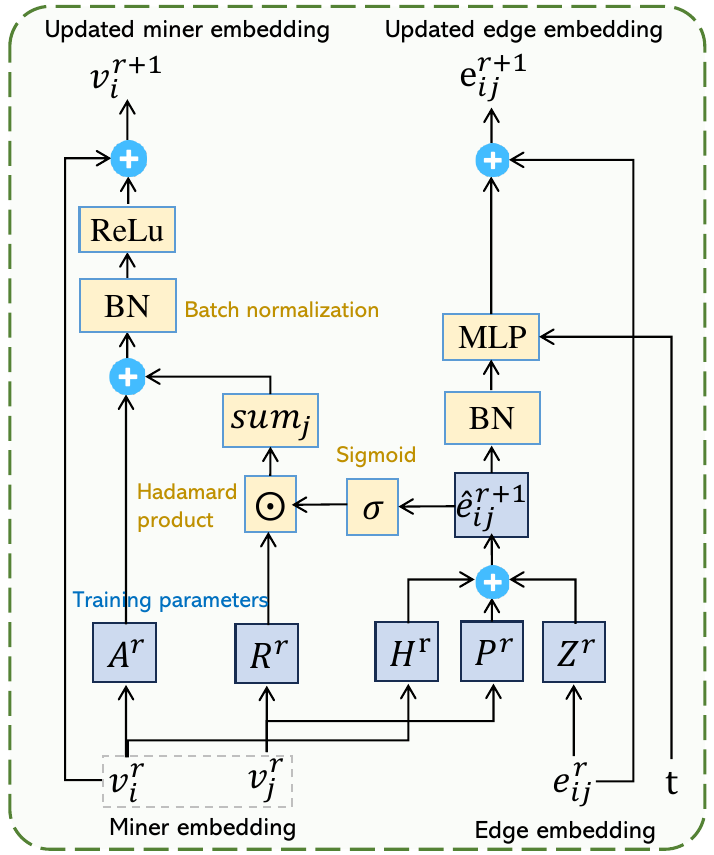}
    \caption{The architecture of the gated graph neural network (GatedGNN) for the graph Resfusion model.}
    \label{gnn}
\end{figure}

Through this approach, we change from estimating noise in normal denoising diffusion models to estimating the residual noise between the degraded input block propagation trajectory and the optimal propagation trajectory. Finally, the loss function of the Resfusion process is given by
\begin{equation}
\mathcal{L}=||res\epsilon-res\epsilon_\theta(G_t,p,t)||^{2}.
\end{equation}

The details of training and sampling in the Resfusion process are shown in Algorithm \ref{Traning algorithm} and Algorithm \ref{Sampling algorithm}.

\begin{table}[t]
	\renewcommand{\arraystretch}{1.4} %控制行高
	\caption{ Key Parameters of Communication Channels.}\label{table} \centering %\tabcolsep=5pt
	\begin{tabular}{m{5.0cm}<{\raggedright}|m{2.5cm}<{\centering}} %两列的长度
		\hline		
		\textbf{Parameters} & \textbf{Values}\\		
  	\hline
		Block size $(S_{block})$ & $1$\:$\rm{MB}$ \cite{liao2024graph}  \\
		\hline
		Transmit power of the IoT devices $(\rho)$ & $23$\:$\rm{dBm}$ \cite{lauridsen2018empirical} \\	
		\hline
		Noise power density $(N_0)$ & $-174$\:$\rm{dBm/Hz}$ \cite{nasir2018fractional} \\
		\hline
		Path-loss coefficient $(\varepsilon)$  &  $3.38$ \cite{liao2024graph}  \\	
		\hline		
		Bandwidth between adjacent miners $(W)$ &  $22$\:$\rm{MHz}$  and $100$\:$\rm{MHz}$ \cite{akpakwu2017survey, garcia2021tutorial}\\
		\hline	
            Unit channel gain $(a)$ &  $-30$\:$\rm{dB}$\cite{su2022energy} \\
        \hline	
	\end{tabular}\label{parameter}
\end{table}
\begin{figure}[ht]
    \centering
    \includegraphics[width=0.9\linewidth]{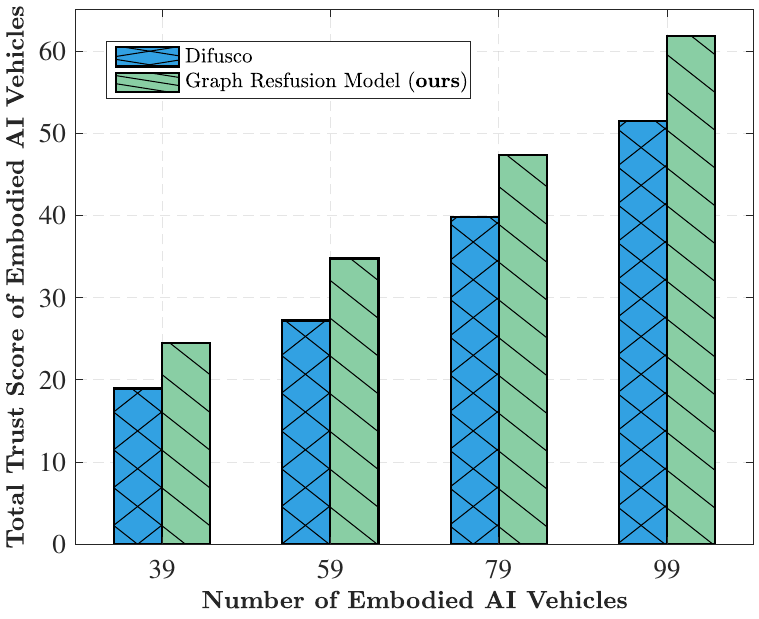}
    \caption{The total trust score of miners corresponding to different numbers of embodied AI vehicles.}
    \label{trust}
\end{figure}

% Consortium Blockchain 
\subsection{Gated Graph Neural Network (GatedGNN) for Resfusion}\label{GNN}
To better estimate $res\epsilon$, we propose a GatedGNN to train the node and edge embeddings for the Resfusion reverse process. In consortium blockchain-enabled MEANETs, the miner network is represented as a graph model. Meanwhile, GatedGNNs are designed to generate embeddings for nodes and edges, unlike traditional GNNs such as graph convolutional networks or graph attention networks\cite{liao2024graph}, which primarily focus on node embeddings. Moreover, the capability that concurrently considers nodes and edges makes GatedGNNs particularly advantageous for tasks that require edge feature prediction, such as finding an optimal block propagation trajectory. Therefore, it is a rational and well-considered choice to apply the GatedGNN for Resfusion.   

\begin{figure*}[t]
    \centering
    \begin{subfigure}[b]{0.40\textwidth}
        \centering
        \includegraphics[width=0.99\textwidth]{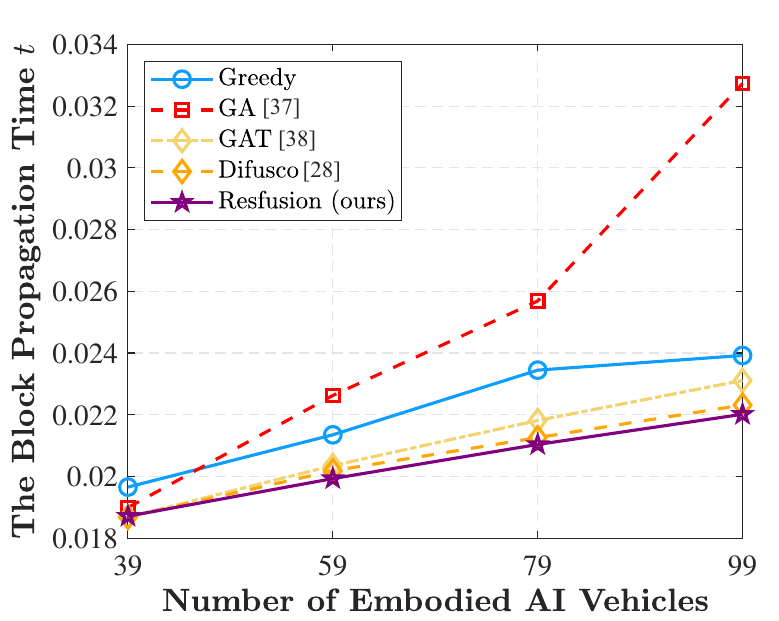}
        \caption{Channel bandwidth $W=22\:\rm{MHz}$.}
    \end{subfigure}
    \begin{subfigure}[b]{0.40\textwidth}
        \centering
        \includegraphics[width=0.99\textwidth]{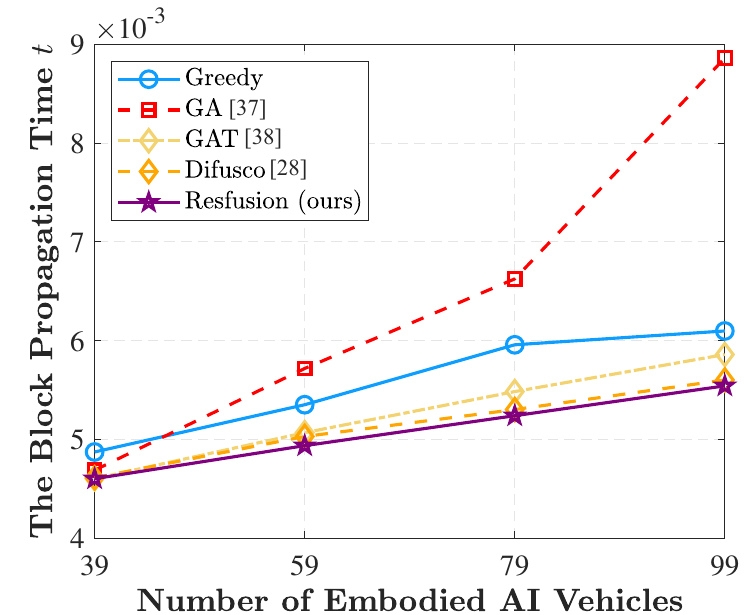}
        \caption{Channel bandwidth $W=100\:\rm{MHz}$.}
    \end{subfigure}
    \caption{The optimal block propagation time $t$ corresponding to different numbers of embodied AI vehicles in different channel bandwidths.}
    \label{t}
\end{figure*}

As shown in Fig. \ref{gnn}, the GatedGNN introduces edge gates, residual connections, and batch normalization to create an anisotropic variant of GNNs. Specifically, the components of the GatedGNN enable it to explicitly update edge features alongside node features, enhancing its flexibility and effectiveness in capturing complex relationships within the miner graph. Therefore, the update process of the edge embeddings in the GatedGNNs is given by\cite{dwivedi2023benchmarking}
\begin{align}
\hat{\boldsymbol{e}}_{ij}^{r+1} &= \boldsymbol{H}^r \boldsymbol{e}_{ij}^r + \boldsymbol{P}^r \boldsymbol{v}_i^r + \boldsymbol{Z}^r \boldsymbol{v}_j^r,\label{e^} \\
\boldsymbol{e}_{ij}^{r+1} &= \boldsymbol{e}_{ij}^{r} + \mathrm{MLP}_{e}(\mathrm{BN}(\hat{\boldsymbol{e}}_{ij}^{r+1})) + \mathrm{MLP}_{t}(\boldsymbol{{t}}),\label{e} 
\end{align}
where $\boldsymbol{v}_i^r$ represents the embedding of miner node $i$ in the GatedGNN layer $r\in\{1,\ldots, N\}$, and $\boldsymbol{e}_{ij}^r$ denotes the embedding of the edge associated with the $i$-th miner in layer $r$. The matrices \( \boldsymbol{H}^{r}, \boldsymbol{P}^{r}\), and \(\boldsymbol{Z}^{r} \in \mathbb{R}^{d \times d} \) represent learnable parameters in layer $r$, and $d$ denotes the pre-configured dimension of the matrices. $\mathrm{BN}(\cdot)$ denotes the batch normalization operator, and $\mathrm{MLP}(\cdot)$ indicates a $2$-layer multi-layer perceptron. 

Based on (\ref{e^}) and (\ref{e}), the update of the miner node embedding in the layer $r+1$ can be given by
\begin{equation}
\boldsymbol{v}_{i}^{r+1} = \boldsymbol{v}_i^r + \mathrm{ReLu}(\mathrm{BN}(\boldsymbol{A}^r \boldsymbol{v}_i^r + \sum_{j\in \mathcal{N}_i}(\sigma(\hat{\boldsymbol{e}}_{ij}^{r+1}) \odot \boldsymbol{R}^r \boldsymbol{v}_j^r))),
\end{equation}
where $\mathrm{ReLu}(\cdot)$ denotes the ReLU activation, \( \sigma \) represents the sigmoid function, \( \odot \) represents the Hadamard product, \( \mathcal{N}_i \) denotes the neighborhoods of miner \( i \), and $\boldsymbol{A}^r$,  $\boldsymbol{R}^{r}$ $\in \mathbb{R}^{d \times d}$ are learnable parameters in layer $r$.

In the initialization process, $\boldsymbol{e}_{ij}^0$ are assigned the corresponding values from $G_t$ while miner node features $\boldsymbol{v}_i^0$ are derived from the miner node coordinates $p$ and initialized utilizing sinusoidal encoding to effectively capture the characteristics of miner nodes.

\section{Simulation Results}\label{results}
\begin{figure*}
    \begin{subfigure}[b]{0.24\textwidth}
        \centering
        \includegraphics[width=0.99\textwidth]{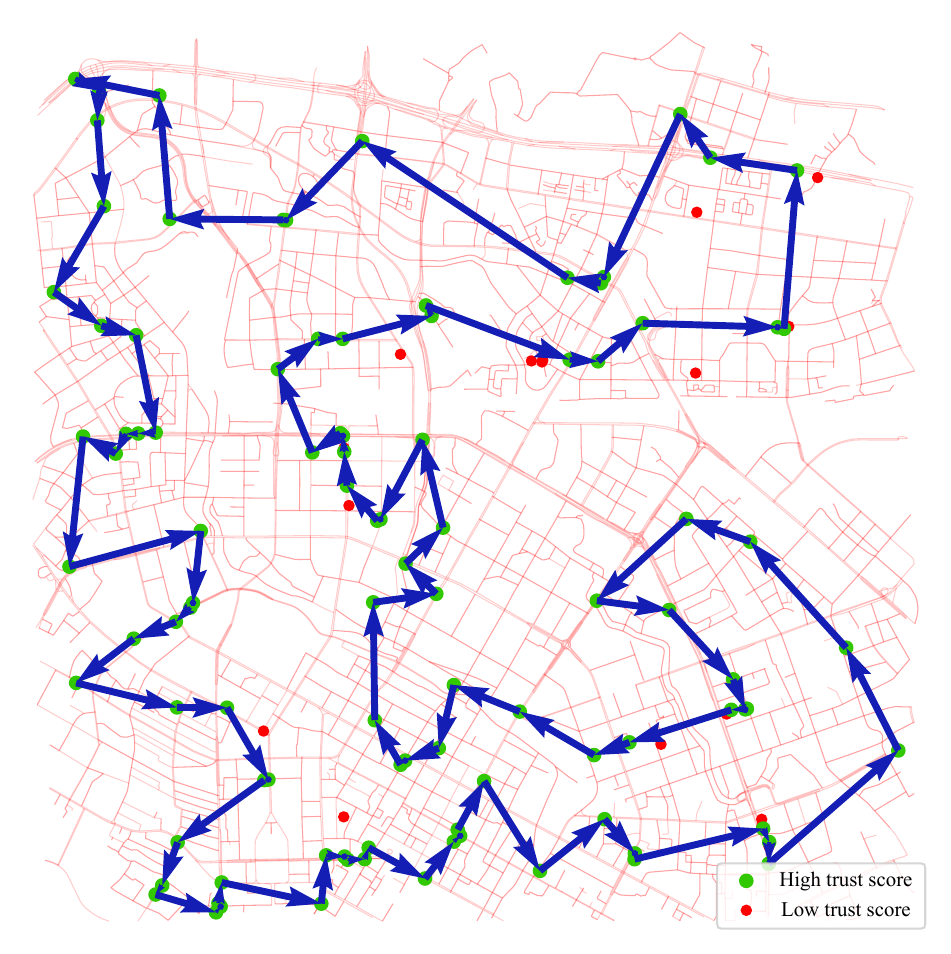}
        \caption{At time $t$.}
        \label{chengdu1}
    \end{subfigure}
    %\hspace{0.5cm} % 调整子图之间的水平间距
    \begin{subfigure}[b]{0.24\textwidth}
        \centering
        \includegraphics[width=1\textwidth]{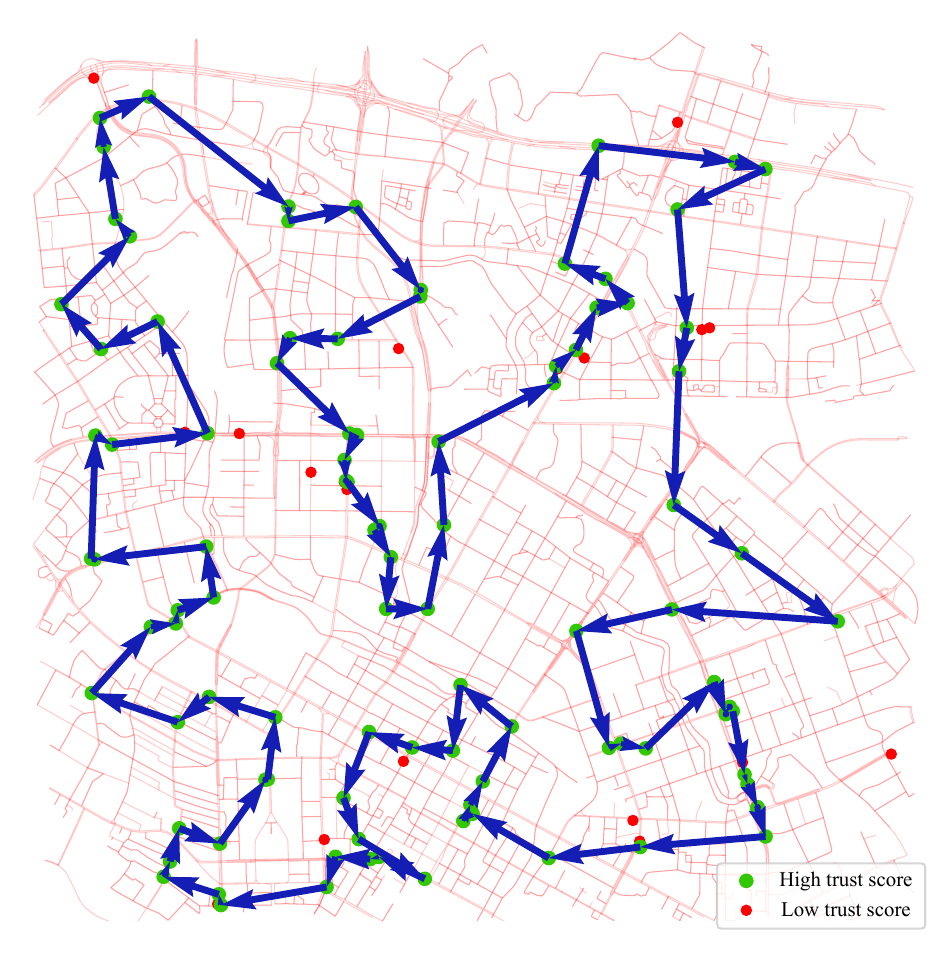}
        \caption{At time $t+30\:\rm{s}$.}
        \label{chengdu2}
    \end{subfigure}
    \begin{subfigure}[b]{0.24\textwidth}
        \centering
        \includegraphics[width=1\textwidth]{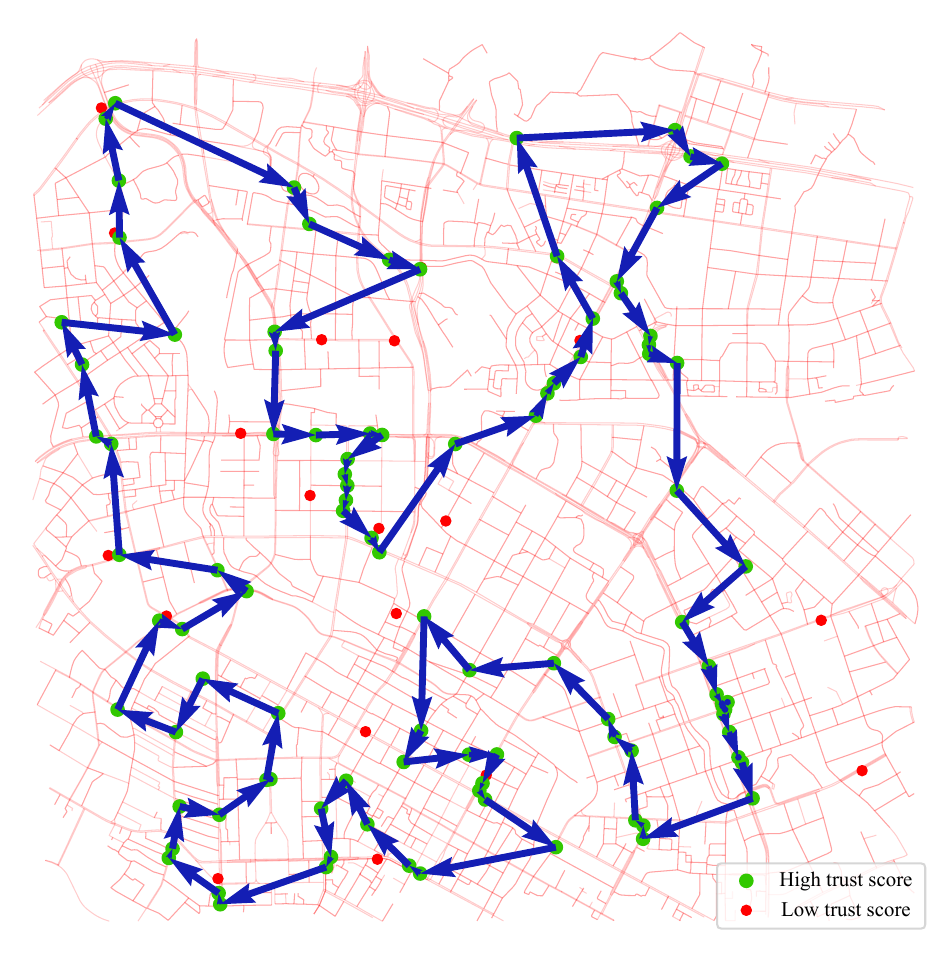}
        \caption{At time $t+60\:\rm{s}$.}
        \label{chengdu3}
    \end{subfigure}
    \begin{subfigure}[b]{0.24\textwidth}
        \centering
        \includegraphics[width=1\textwidth]{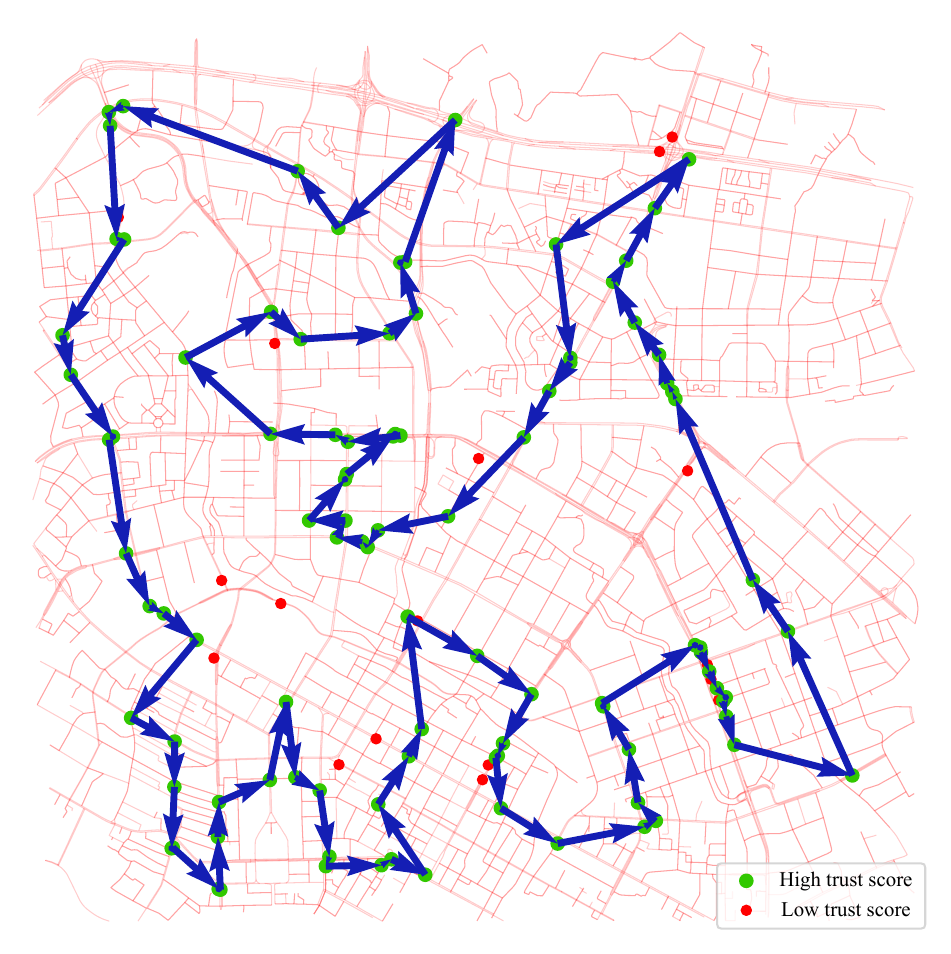}
        \caption{At time $t+90\:\rm{s}$.}
        \label{chengdu4}
    \end{subfigure}
    
    \begin{subfigure}[b]{0.24\textwidth}
        \centering
        \includegraphics[width=0.99\textwidth]{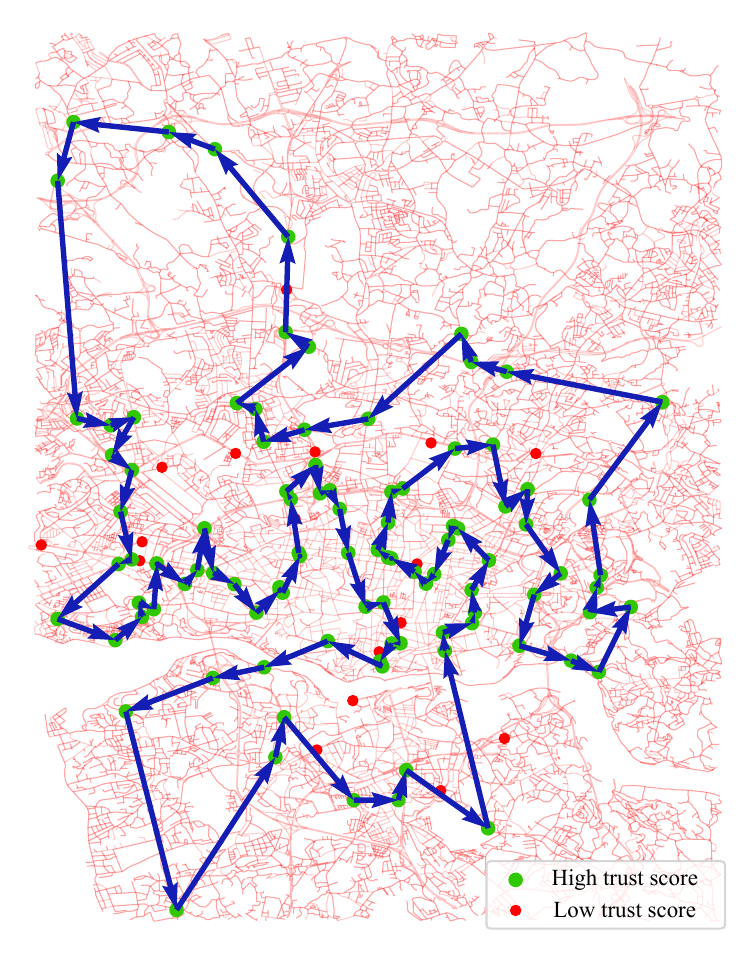}
        \caption{At time $t$.}
        \label{porto1}
    \end{subfigure}
    \begin{subfigure}[b]{0.24\textwidth}
        \centering
        \includegraphics[width=0.99\textwidth]{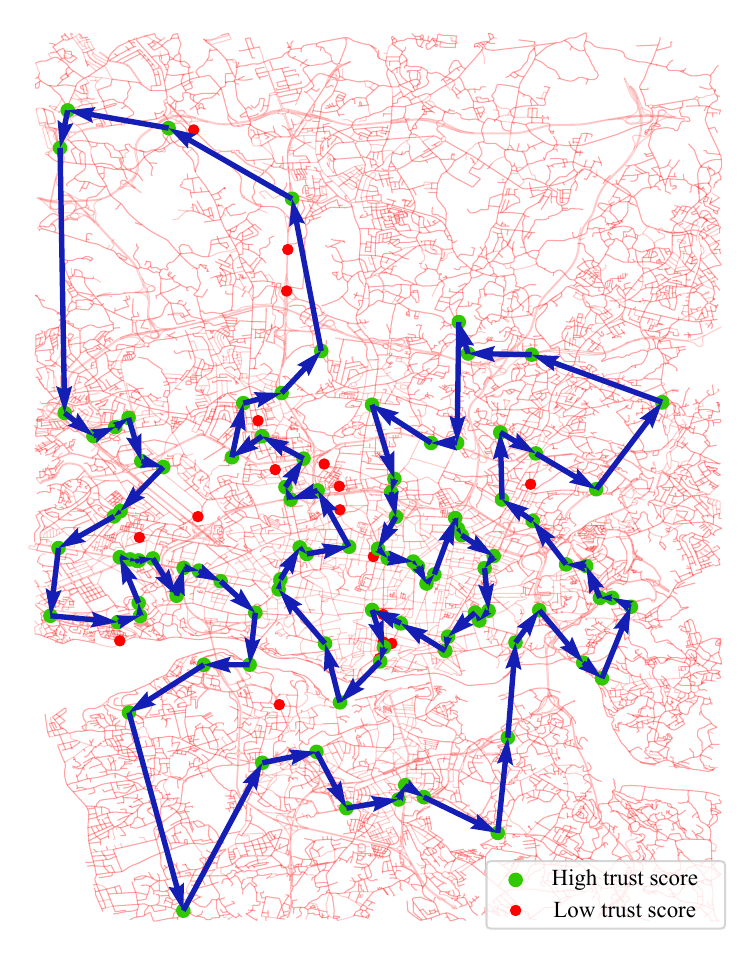}
        \caption{At time $t+30\:\rm{s}$.}
        \label{porto2}
    \end{subfigure}
    \begin{subfigure}[b]{0.24\textwidth}
        \centering
        \includegraphics[width=0.99\textwidth]{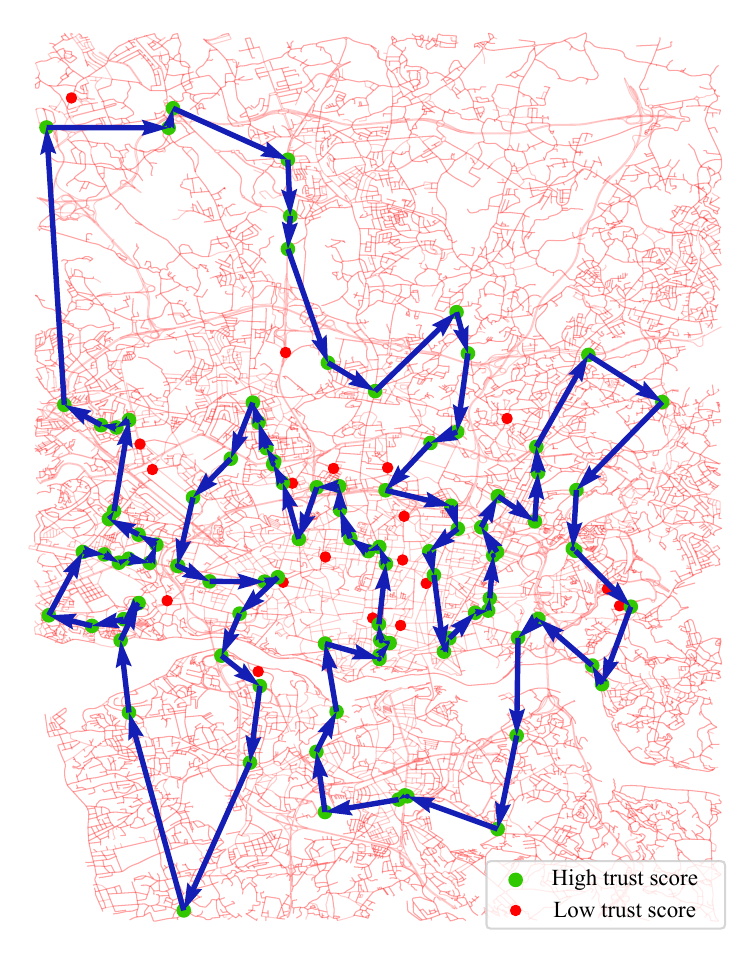}
        \caption{At time $t+60\:\rm{s}$.}
        \label{porto3}
    \end{subfigure}
    \begin{subfigure}[b]{0.24\textwidth}
        \centering
        \includegraphics[width=0.99\textwidth]{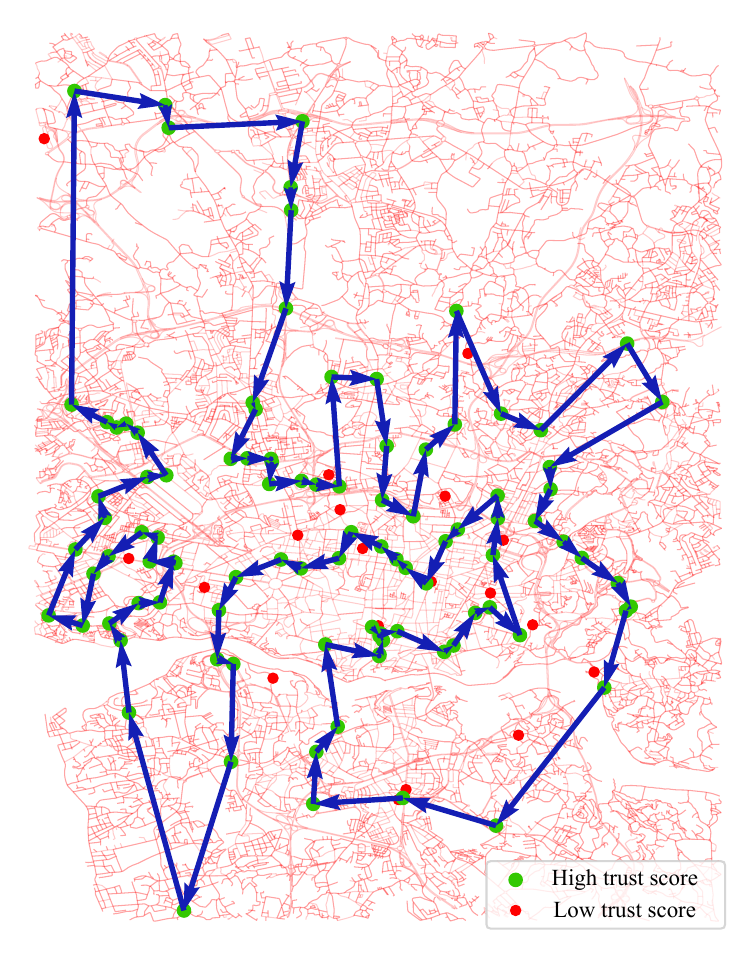}
        \caption{At time $t+90\:\rm{s}$.}
        \label{porto4}
    \end{subfigure}
    \caption{The optimal block propagation trajectories corresponding to different times and embodied AI vehicle datasets. Figs. \ref{chengdu1}, \ref{chengdu2}, \ref{chengdu3}, and \ref{chengdu4} show the block propagation trajectories in Chengdu city at different times with a time slot $30 \: s$, while Figs. \ref{porto1}, \ref{porto2}, \ref{porto3}, and \ref{porto4} show the block propagation trajectories in Porto city at different times with a time slot $30 \: s$.}
    \label{propagation_figure}
\end{figure*}
\begin{figure*}
    \begin{subfigure}[b]{0.24\textwidth}
        \centering
        \includegraphics[width=0.99\textwidth]{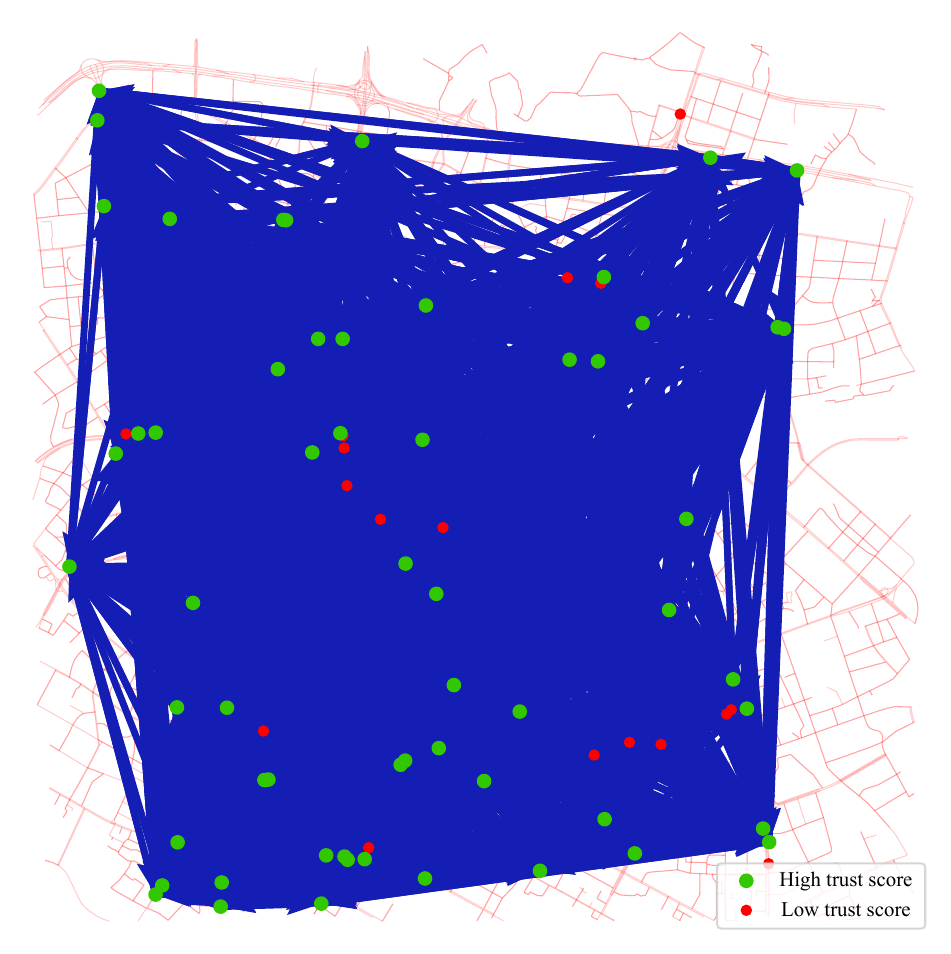}
        \caption{At Resfusion step $=0/368$.}
        \label{step0}
    \end{subfigure}
    %\hspace{0.5cm} % 调整子图之间的水平间距
    \begin{subfigure}[b]{0.24\textwidth}
        \centering
        \includegraphics[width=1\textwidth]{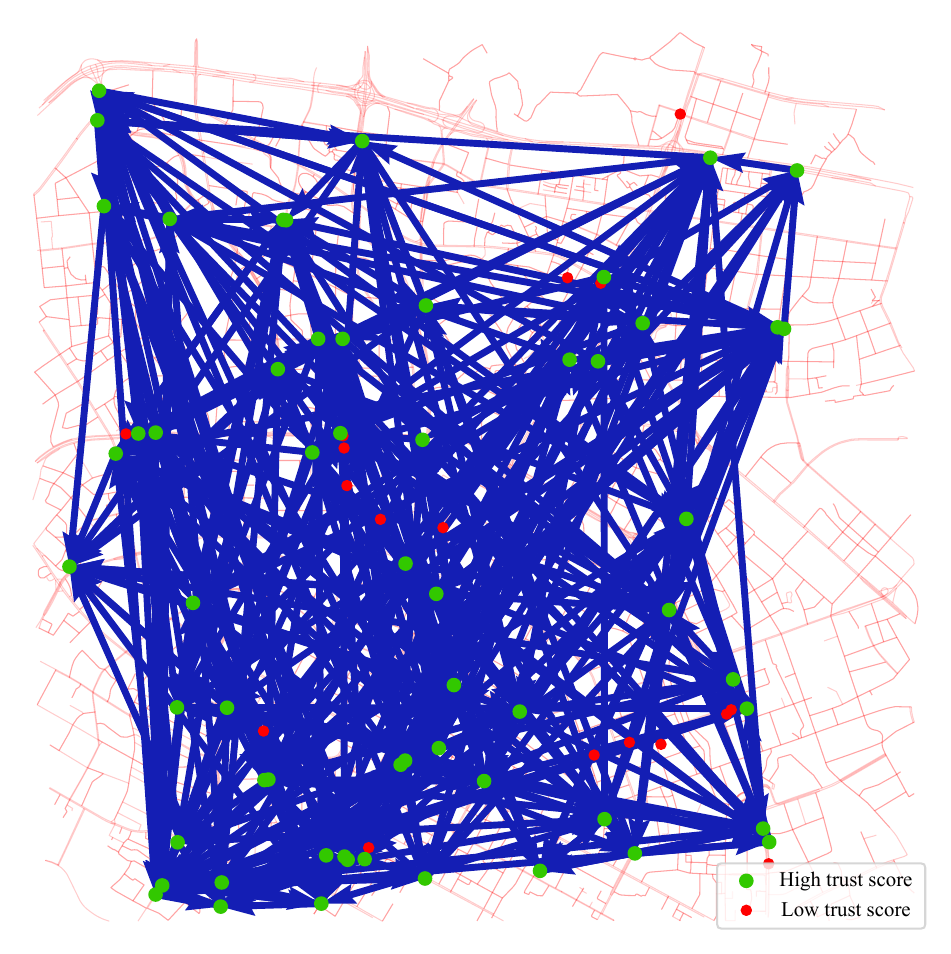}
        \caption{At Resfusion step $=320/368$.}
        \label{step320}
    \end{subfigure}
    \begin{subfigure}[b]{0.24\textwidth}
        \centering
        \includegraphics[width=1\textwidth]{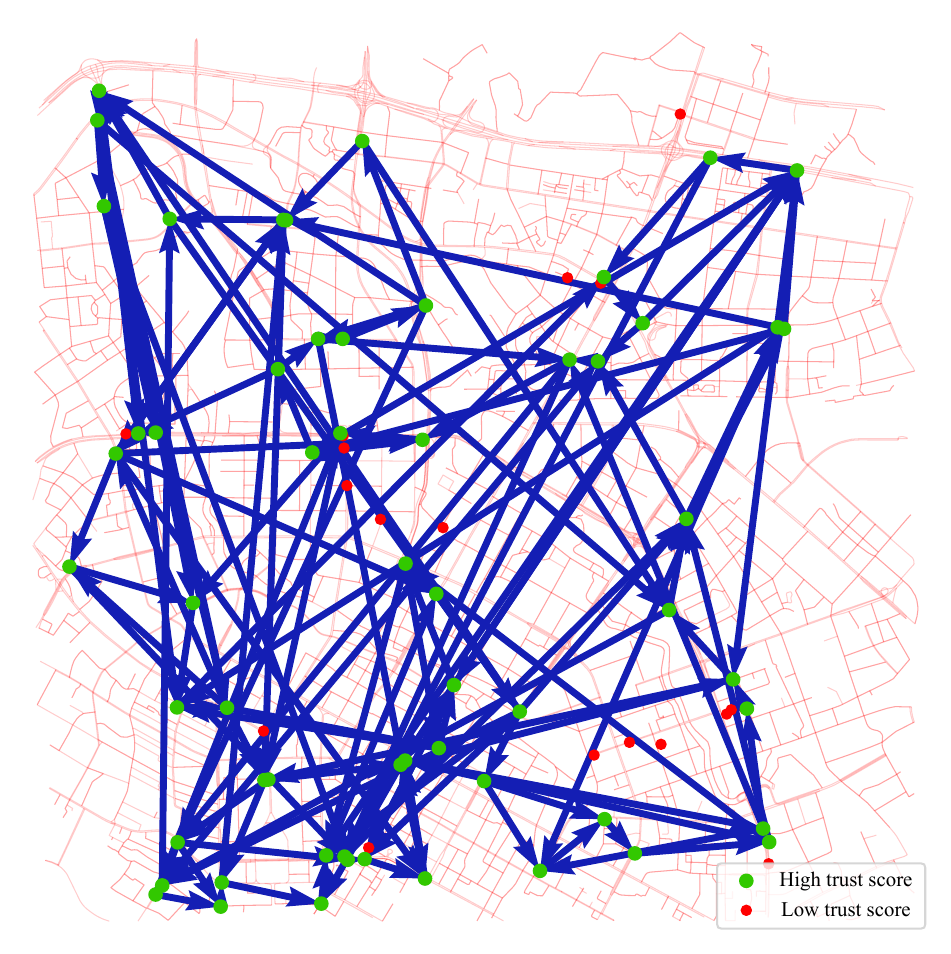}
        \caption{At Resfusion step $=338/368$.}
        \label{step338}
    \end{subfigure}
    \begin{subfigure}[b]{0.24\textwidth}
        \centering
        \includegraphics[width=1\textwidth]{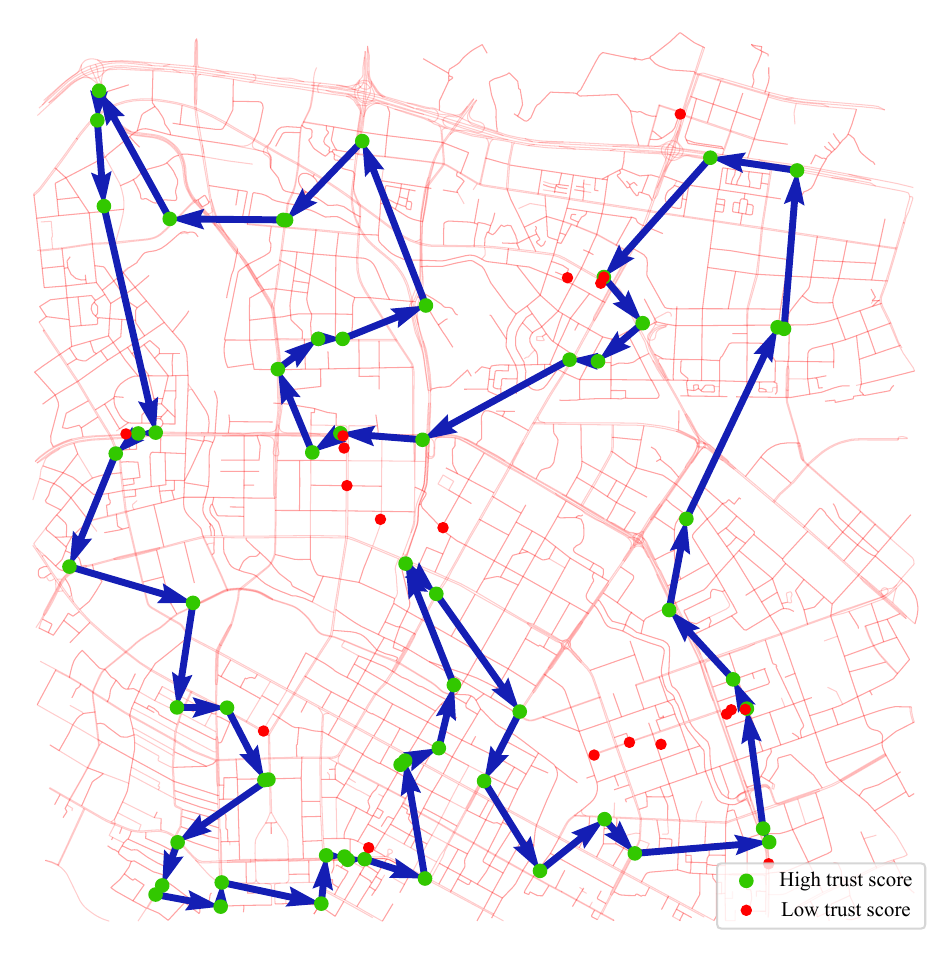}
        \caption{At Resfusion step $=368/368$.}
        \label{step368}
    \end{subfigure}
    \caption{Specific Resfusion processes at different steps.}
    \label{steps}
\end{figure*}
In this section, we compare the proposed graph Resfusion model with other typical routing mechanisms: i) \textit{Greedy mechanism}. In the simulation setup, the greedy mechanism consistently evaluates the distances between miners and prioritizes selecting the nearest adjacent miners for block propagation; ii) \textit{Genetic algorithm}. The genetic algorithm (GA) \cite{mirjalili2019genetic} is a classic heuristic algorithm, which is widely utilized for optimization problems; iii) \textit{Graph attention network}. The GAT in \cite{kool2018attention} has a strong graph processing capability, which is suitable for the block propagation problem in consortium blockchain-enabled MEANETs; iv) \textit{Difusco}. The Difusco is proposed in \cite{sun2023difusco}, which utilizes a graph diffusion model to solve NP-complete problems. Similar to \cite{sun2023difusco}, we decode the heatmap scores generated by Difusco and the graph Resfusion model using a $2$-opt algorithm.

\subsection{Implementation of Simulations}
%说一下数据集来自哪里？
\subsubsection{Parameter Setting}
For the simulation parameter settings, we apply the threshold of trust score $\lambda=0.5$ to constrain the low-trust score miner nodes and the number of untrusted miners $(V-K)=19$ to remove the untrusted miner nodes in the miner network. Without loss of generality, we set the parameters of the cloud according to \cite{liao2024graph, kang2018blockchain}. Moreover, in the GatedGNN model designed for the Resfusion, we employ a $12$-layer GatedGNN with a pre-configured matrix dimension of $d=256$ for the width. Note that we conduct the simulations on NVIDIA GeForce RTX 3080Ti, and the key parameters are presented in Table \ref{parameter}\cite{liao2024graph,  akpakwu2017survey, garcia2021tutorial, lauridsen2018empirical, nasir2018fractional}. 
\subsubsection{Dataset and Scenario Setting}
Following \cite{sun2023difusco}, we use the Concorde exact solver to generate and label training data to train the graph Resfusion model. Subsequently, we conduct simulations using the well-trained graph Resfusion model on two real-world vehicle trajectory datasets: the Chengdu dataset and the Porto dataset. The Chengdu dataset contains $3,493,918$ vehicle trajectories with a sampling interval of $3\:\rm{s}$. The Porto dataset includes $1,189,730$ vehicle trajectories with a sampling interval of $15\:\rm{s}$. We consider these real-world vehicles to be embodied AI vehicles driven by lightweight large language models\cite{zhang2025embodied}. The communication between these embodied AI vehicles is facilitated through orthogonal frequency division multiple access. Moreover, every embodied AI vehicle can serve as a miner in the consortium blockchain-enabled MEANETs. For each embodied AI vehicle trajectory dataset, we randomly select the required number of trajectories $V=(58, 78, 98, 118)$ from numerous trajectories and conduct simulations on the embodied AI vehicle positions at the same time slot for these trajectories.

\subsection{Trustworthiness and Efficiency}
Figure \ref{trust} shows the total trust scores of embodied AI vehicles corresponding to different numbers of embodied AI vehicles. As demonstrated, the proposed graph Resfusion model integrated with the trust calculation mechanism achieves a significantly higher total trust score for embodied AI vehicles compared to the graph Resfusion model without the trust calculation mechanism. This improvement can be attributed to the cloud-based trust calculation mechanism, which effectively filters out embodied AI vehicles with low trust scores, thus improving the overall reliability and trustworthiness of block propagation in MEANETs. Specifically, the trust score for $99$ embodied AI vehicles increases by an impressive $19.60\%$ when the trust calculation mechanism is employed, highlighting its effectiveness in fostering a more secure and robust network environment. By ensuring that only embodied AI vehicles with high trust scores participate in block propagation, the mechanism not only reduces the likelihood of malicious activities but also improves the efficiency and integrity of data dissemination in consortium blockchain-enabled MEANETs.

Figure \ref{t} depicts the optimal block propagation time corresponding to different numbers of embodied AI vehicles in different channel bandwidths, using OFDMA. Considering different communication environments, we conduct simulations in bandwidths of $W=22\:\rm{MHz}$\cite{akpakwu2017survey} and $W=100\:\rm{MHz}$\cite{garcia2021tutorial}, corresponding to WiFi network and 5G network, respectively. As shown in Fig. \ref{t}, the proposed graph Resfusion model outperforms the other four algorithms, with the performance gap widening as the number of embodied AI vehicles increases, thus reflecting on the effectiveness of the graph Resfuion model. As we can observe, the performance ranking of the algorithms is as follows: the proposed graph Resfusion model, Difusco, GAT, GA, and Greedy. Specifically, as the number of embodied AI vehicles increases, the block propagation time under the GA mechanism grows at the fastest rate due to its limited ability to efficiently handle a larger number of nodes. In contrast, GAT, Difusco, and graph Resfusion demonstrate superior performance, benefiting from the integration of graph network mechanisms. Moreover, the results confirm that combining graph structures with diffusion processes can enhance performance. Furthermore, graph Resfusion surpasses Difusco, demonstrating that the incorporation of degraded block propagation trajectories introduces valuable prior knowledge, which helps graph Resfusion to reduce the sampling space and enhance the performance in optimizing block propagation trajectories.

\subsection{Strong Dynamic Adaptability}
Figure \ref{propagation_figure} shows the optimal block propagation trajectories produced by the graph Resfusion model at different times across different embodied AI vehicle trajectory datasets. 
Specifically, we conduct two sets of experiments in two different embodied AI vehicle datasets with the embodied AI vehicle number $V=118$. Figs. \ref{chengdu1}, \ref{chengdu2}, \ref{chengdu3}, and \ref{chengdu4} show the block propagation trajectories in Chengdu city, while Figs. \ref{porto1}, \ref{porto2}, \ref{porto3}, and \ref{porto4} show the block propagation trajectories in Porto city. For better presentation, red points are used to represent the embodied AI vehicles that have unacceptably low trust scores, and green points are used to denote the embodied AI vehicles that have high trust scores, where the trust scores are calculated by the proposed trust calculation mechanism in Section \ref{cloud1}. Moreover, the figure background is the city road network and the blue arrows indicate the block propagation trajectories. As observed, the proposed graph Resfusion model effectively generates optimal block propagation trajectories across different times and cities, which have no unreasonable costs in length and are clearly organized for efficiency, thus reflecting on the strong dynamic adaptability of the graph Resfusion model. This capability is exactly suitable for the rapidly changing scenario of MEANETs. Furthermore, it is intuitive to observe that the embodied AI vehicles that have low trust scores are perfectly excluded to ensure trustworthiness in the block propagation process.

\subsection{Visualization of the Denoising Process}
In Fig. \ref{steps}, we illustrate the specific steps of the denoising process of the proposed graph Resfusion model. According to (\ref{get T'}), we only need to train the
forward process and execute the reverse process from $0$ to $T^{\prime}$ profiting from the introduction of residuals, as depicted in Fig. \ref{steps}, at the representative steps $0$, $320$, $338$, and $368$. Specifically, the Resfusion process achieves an optimal block propagation trajectory from the degraded block propagation trajectory within just only $368$ steps, compared to the $1000$ steps required in a standard diffusion process, which makes a significant contribution to reducing the inference time for the denoising process. 

\begin{figure}
    \begin{subfigure}[b]{0.24\textwidth}
        \centering
        \includegraphics[width=0.99\textwidth]{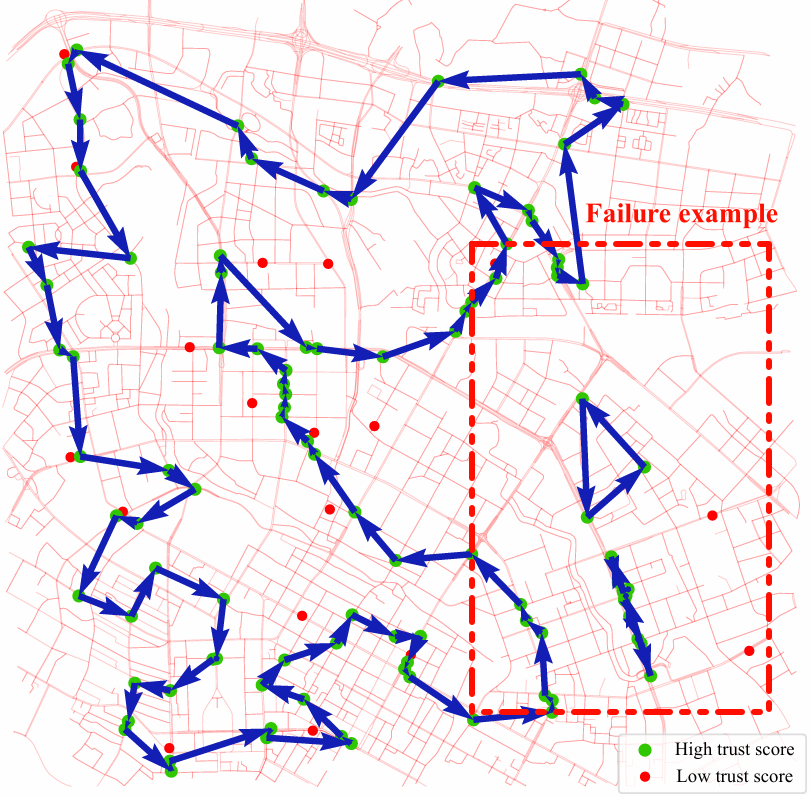}
        \caption{Failure example.}
        \label{wrong}
    \end{subfigure}
    \begin{subfigure}[b]{0.24\textwidth}
        \centering
        \includegraphics[width=1\textwidth]{path_at_different_time/traj_chengdu35_withroad.pdf}
        \caption{Successful example.}
        \label{success}
    \end{subfigure}
    \caption{Examples of failures and successes.}
    \label{wrong_analysis}
\end{figure}
\begin{figure}
    \centering
    \includegraphics[width=0.9\linewidth]{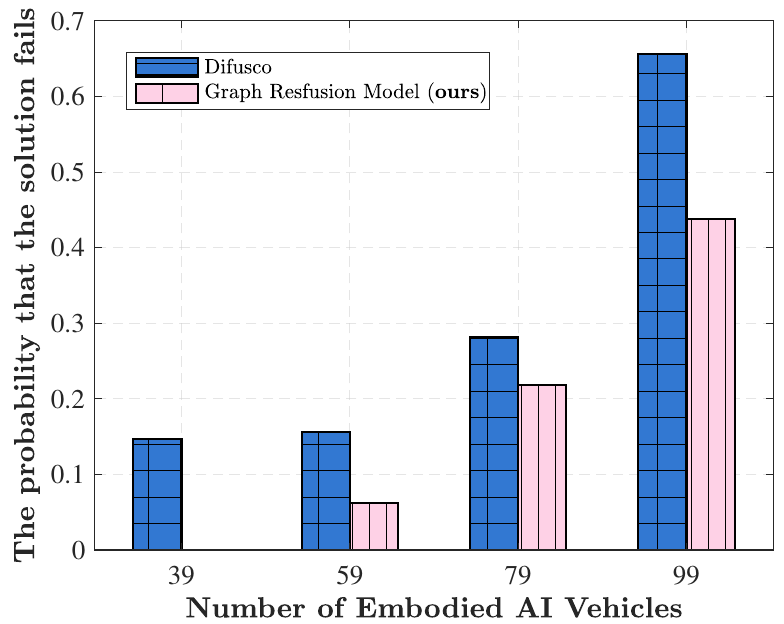}
    \caption{The probability that the solution fails corresponding to different numbers of embodied AI vehicles.}
    \label{probability}
\end{figure}

\subsection{Inference Accuracy and Generalization Performance}
Since the diffusion model covers a huge sampling space during the sampling process, it may result in many block propagation trajectories that fail to meet the requirements of the block propagation mechanism. Figure  \ref{wrong_analysis} presents examples of both failure and success. To ensure the efficiency of block propagation, we aim to minimize the probability of failures in the sampled block propagation trajectory. To evaluate graph Resfusion's capability in reducing failed block propagation trajectories compared to Difusco, we perform $32$ parallel samplings on the same instance and calculated the failure probabilities for both methods. Figure \ref{probability} shows the probability of failure for different embodied AI vehicle numbers. As can be observed, the Difusco model exhibits a higher probability of failure compared to the proposed graph Resfusion model, with the failure probability increasing as the number of embodied AI vehicles increases. Notably, for $39$ embodied AI vehicles, graph Resfusion does not generate any failed block propagation trajectories. For $99$ embodied AI vehicles, the failure probability in the graph Resfusion model is $33.3\%$ lower than that of Difusco. This indicates that the introduction of degraded block propagation trajectories in graph Resfusion provides valuable prior knowledge guidance, which significantly reduces the probability of generating failed block propagation trajectories and highlights the inference accuracy and effectiveness of the graph Resfusion model.

\begin{figure}
    \centering
    \includegraphics[width=0.9\linewidth]{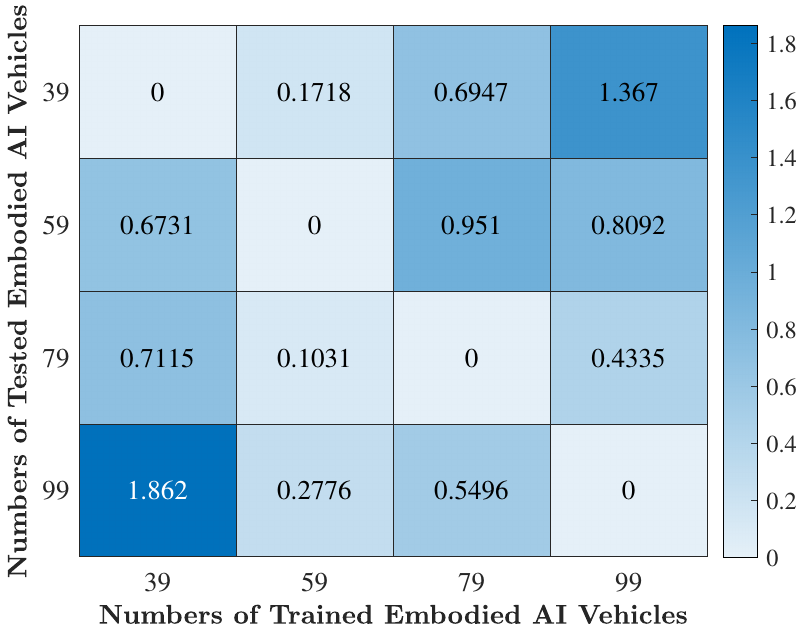}
    \caption{The generalization performance tests corresponding to different numbers of embodied AI vehicles for the graph Resfusion model.}
    \label{heatmap}
\end{figure}

%讲一下热力图里的数据是怎么处理的
In Fig. \ref{heatmap}, we depict the generalization performance tests corresponding to different numbers of embodied AI vehicles for the graph Resfusion model. Specifically, we conduct experiments in a unit square area and calculate the overall lengths of the block propagation trajectories to test how well the model generalizes. The diagonal line in the figure represents the scenario where the model is both trained and tested on the same number of embodied AI vehicles, serving as the benchmark for evaluating the model generalization performance across different embodied AI vehicle numbers. Specifically, the difference between the performance of a given embodied AI vehicle number and the baseline is calculated, followed by determining the proportion of this difference relative to the baseline. This proportion is then recorded for analysis and it is evident that a lower percentage indicates superior generalization performance. As we can observe, the maximum percentage does not exceed $2\%$, demonstrating the strong generalization capability of the proposed graph Resfusion model across varying embodied AI vehicle numbers, which may be attributed to the successful and contributing introduction of the degraded solution and the residual network in the graph Resfuion model.
\section{Conclusion}\label{conclusion}
This paper focuses on improving the performance of consortium blockchain-enabled MEANETs, with an emphasis on optimizing block propagation within consortium blockchains. Specifically, we have proposed a trust calculation mechanism based on the cloud model to ensure trustworthiness during block propagation. Moreover, to achieve effectiveness in block propagation, we have established the graph Resfusion model to generate optimal block propagation trajectories. Finally, simulations demonstrate that compared to other routing mechanisms, the proposed model can achieve a higher trust score for miners and the shortest block propagation time, significantly contributing to enhancing both the efficiency and trustworthiness of block propagation in consortium blockchains. Furthermore, the results highlight the strong dynamic adaptability of the proposed model, making it well-suited to the rapidly changing MEANETs.

For future work, we will study advanced algorithms about graph diffusion models, including scalable diffusion models integrated with transformers, to enhance the efficiency and effectiveness of block propagation. Additionally, we will aim towards exploring directed acyclic graph-based approaches or implementing privacy-preserving model training techniques through federated learning in blockchain-enabled MEANETs.
\bibliographystyle{IEEEtran}
\bibliography{ref}

\begin{IEEEbiography}
[{\includegraphics[width=1in,height=1.25in,clip,keepaspectratio]{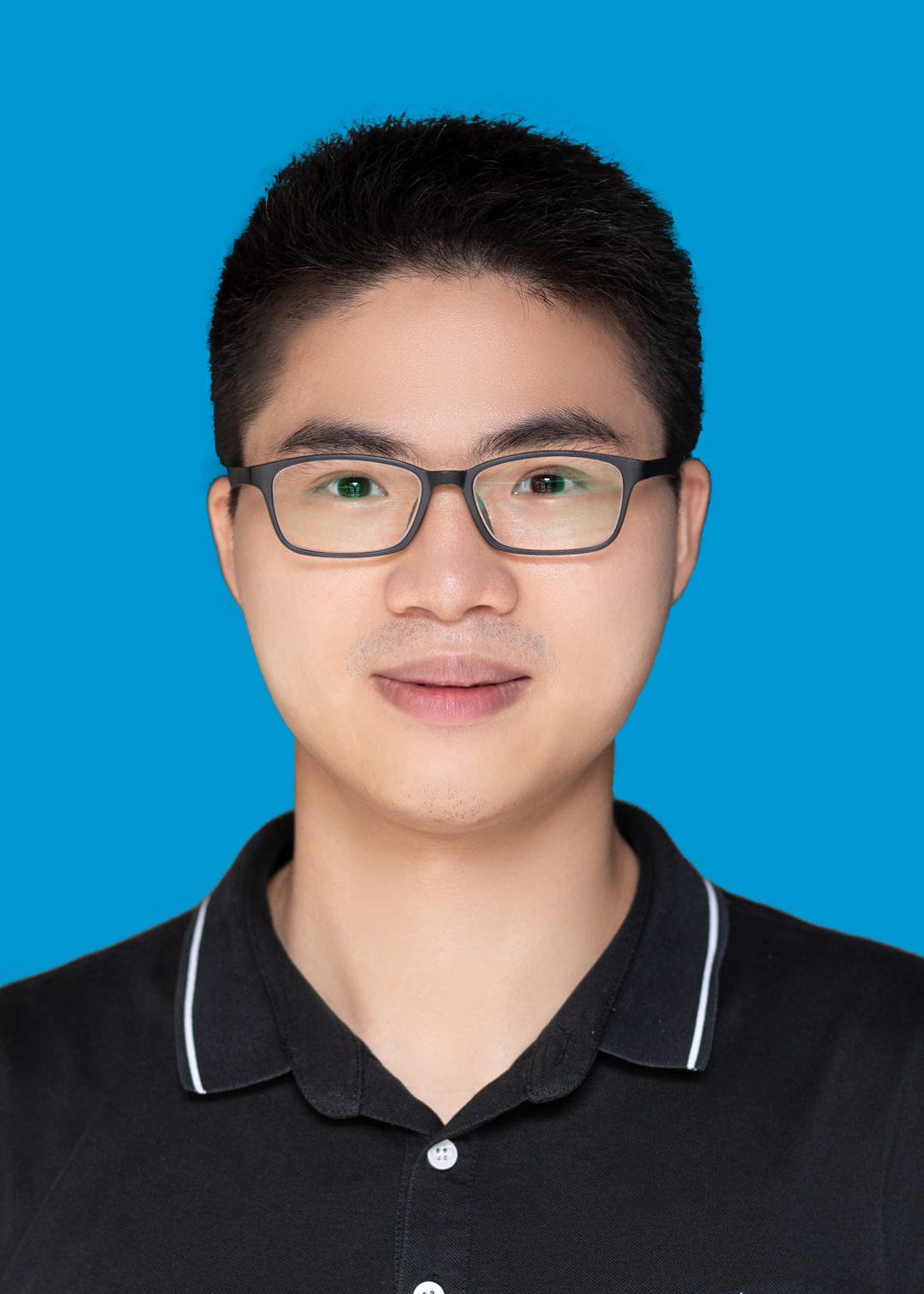}}]{Jiawen Kang} received the Ph.D. degree from Guangdong University of Technology, China, in 2018. He has been a postdoc at Nanyang Technological University, Singapore from 2018 to 2021. He is currently a full professor at Guangdong University of Technology, China. His research interests mainly focus on generative AI, blockchain, security, and privacy protection in wireless communications and networking.
\end{IEEEbiography}

\begin{IEEEbiography}
[{\includegraphics[width=1in,height=1.25in,clip,keepaspectratio]{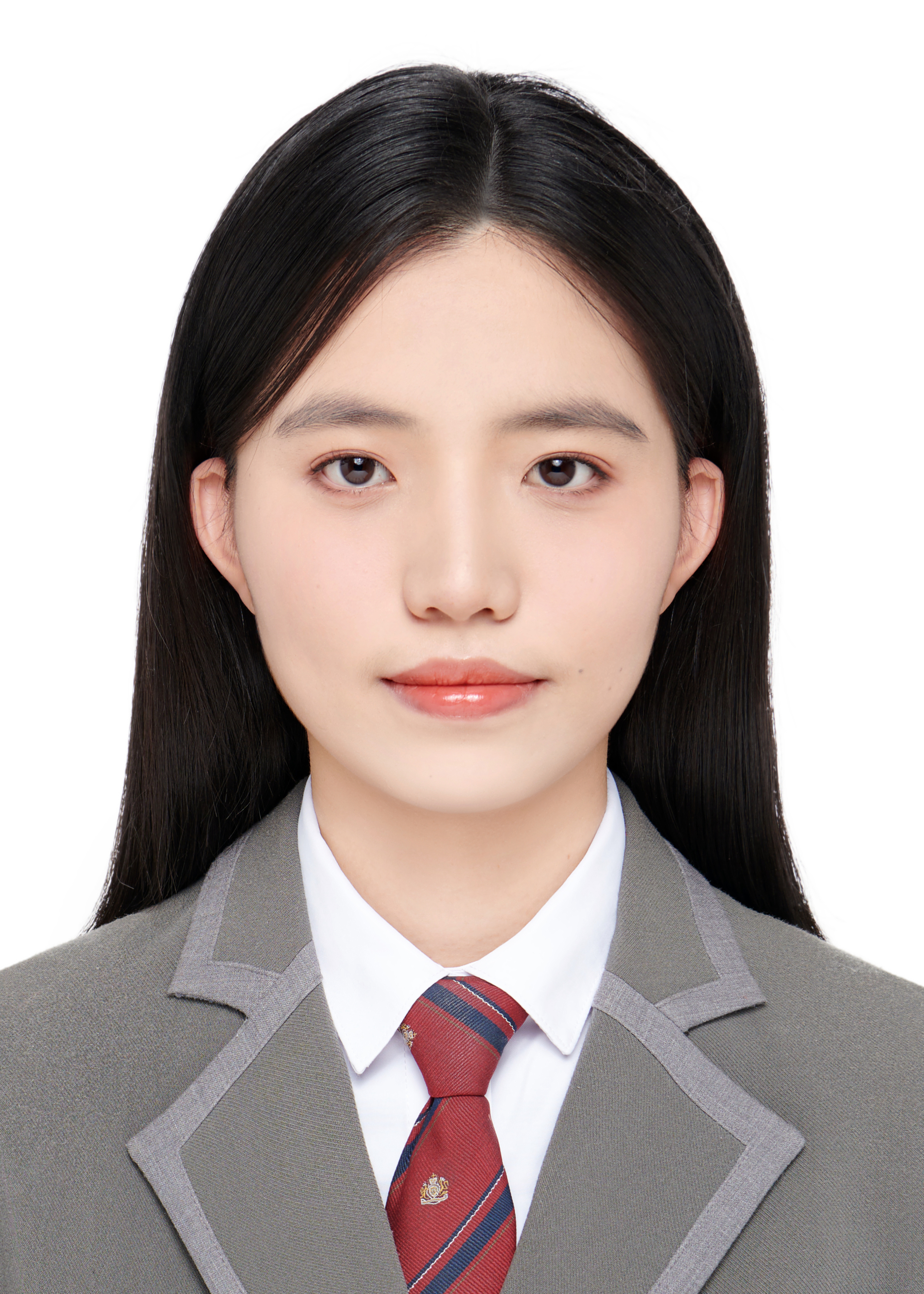}}] 
{Jiana Liao} is currently pursuing a B.Eng. degree from Guangdong University of Technology, Guangzhou, China. She will pursue a Ph.D. degree with the School of Software Engineering, Sun Yat-Sen University, Zhuhai, China. Her research interests mainly focus on graph neural networks, blockchain, and deep reinforcement learning.
\end{IEEEbiography}

\begin{IEEEbiography}
[{\includegraphics[width=1in,height=1.25in,clip,keepaspectratio]{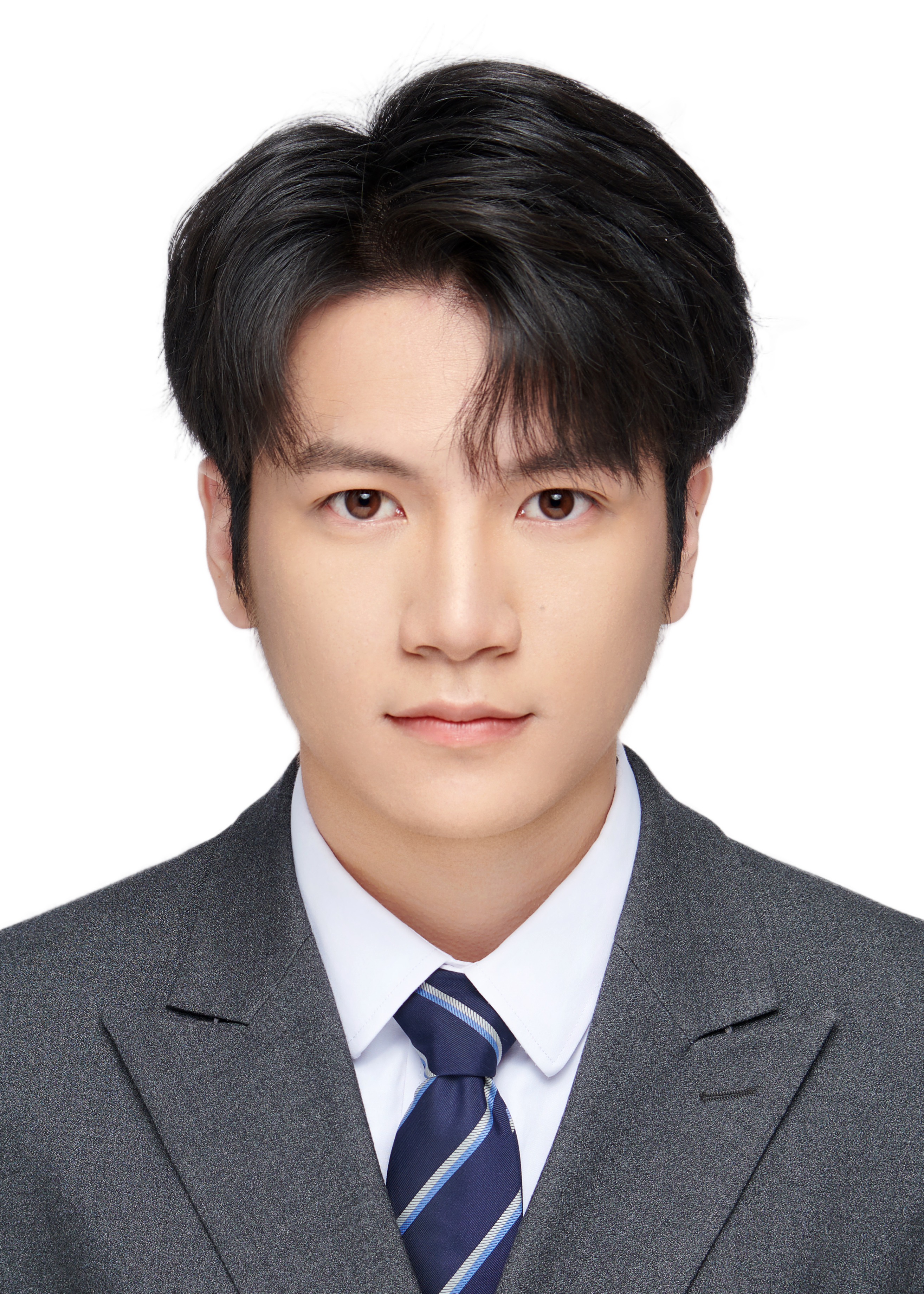}}] 
{Runquan Gao} is currently pursuing a B.Eng. degree from Guangdong University of Technology, Guangzhou, China. He will pursue an M.S. degree with the School of Intelligent Systems Engineering, Sun Yat-Sen University, Shenzhen, China. His research interests mainly focus on edge intelligence, federated learning, and diffusion model.
\end{IEEEbiography}

\begin{IEEEbiography}
[{\includegraphics[width=1in,height=1.25in,clip,keepaspectratio]{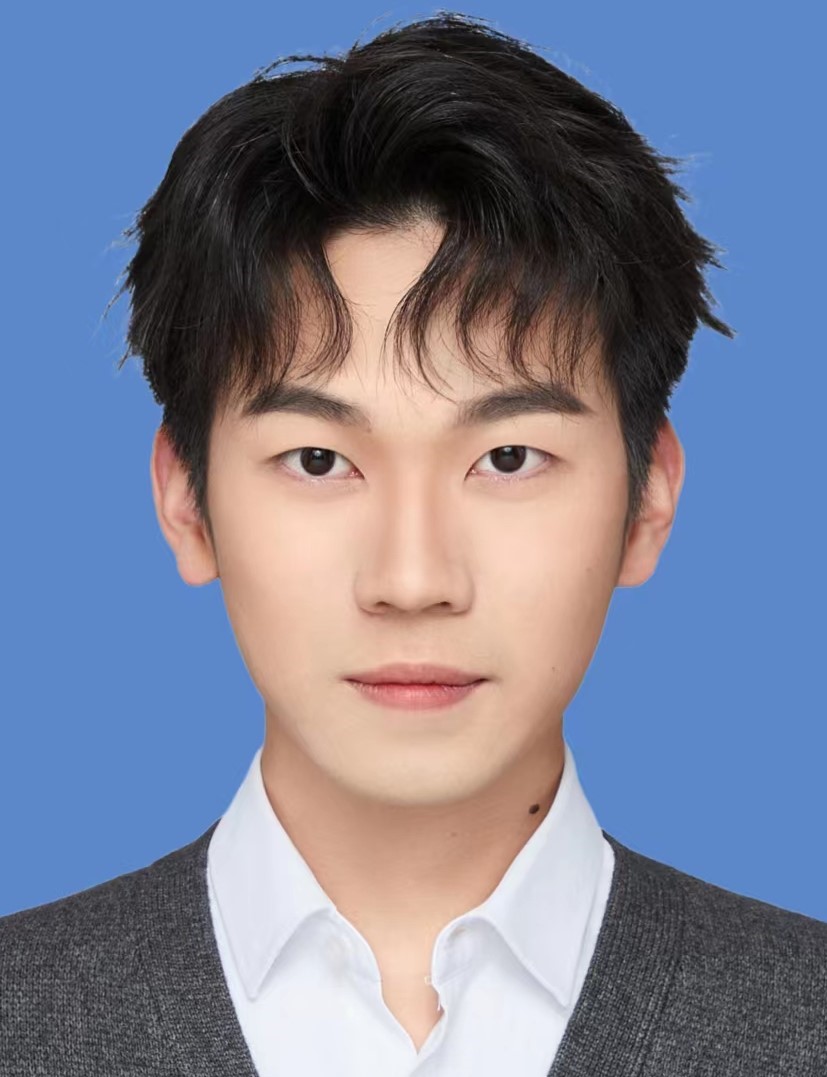}}] 
{Jinbo Wen} received the B.Eng. degree from Guangdong University of Technology, Guangzhou, China, in 2023. He is currently pursuing an M.S. degree with the College of Computer Science and Technology, Nanjing University of Aeronautics and Astronautics, China. His research interests include generative AI, blockchain, edge intelligence, and metaverse.
\end{IEEEbiography}

\begin{IEEEbiography}
[{\includegraphics[width=1in,height=1.25in,clip,keepaspectratio]{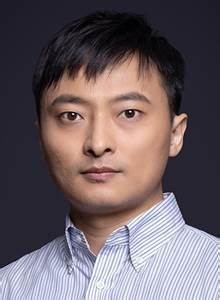}}] 
{Huawei Huang} (Senior Member, IEEE)  received the Ph.D. degree from the University of Aizu, Aizuwakamatsu, Japan, in 2016. He is currently an Associate Professor with Sun Yat-Sen University, Guangzhou, China. He was a Research Fellow of JSPS, and a Program-Specific Assistant Professor with Kyoto University, Kyoto, Japan. His research interests include blockchain and distributed computing. He was the Lead Guest Editor for multiple blockchain special issues at IEEE
 JOURNAL ON SELECTED AREAS IN COMMUNICA
TIONS, and IEEE OPEN JOURNAL OF THE COMPUTER SOCIETY. He also was a TPC Chair for multiple blockchain conferences and workshops.
\end{IEEEbiography}

\begin{IEEEbiography}
[{\includegraphics[width=1in,height=1.25in,clip,keepaspectratio]{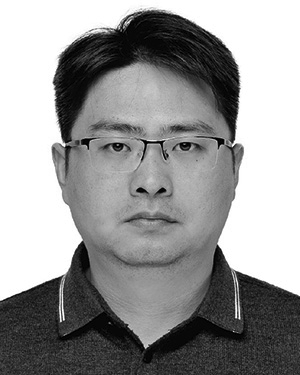}}] 
{Maomao Zhang} received the Ph.D. degree in information and communication engineering with the College of Communication Engineering, Army Engineering University of PLA, Nanjing, China. He is currently a professor with Anhui Engineering Research Center for Agricultural Product Quality Safety Digital Intelligence, Fuyang Normal University. His research interests include deep learning and automatic modulation recognition and their applications in wireless communication network.
\end{IEEEbiography}

\begin{IEEEbiography}
[{\includegraphics[width=1in,height=1.25in,clip,keepaspectratio]{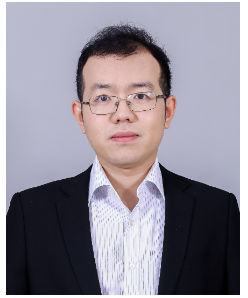}}]
{Changan Yi} received the Ph.D. degree from the University of Manitoba, MB, Canada, in 2018. He is currently a Professor with the College of Computer Science and Technology, Nanjing University of Aeronautics and Astronautics, Nanjing, China. His research interests include game theory, queueing theory, machine learning, and its applications in edge/fog computing, IoT, 5G and beyond.
\end{IEEEbiography}

\begin{IEEEbiography}
[{\includegraphics[width=1in,height=1.25in,clip,keepaspectratio]{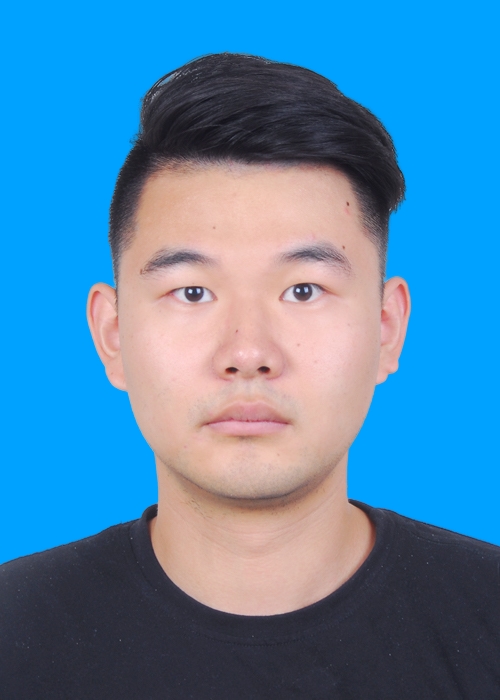}}]
{Tao Zhang} (Member, IEEE) received the B.S degree in Internet of Things engineering from the Beijing University of Posts and Telecommunications
(BUPT) and the Queen Mary University of London in 2018, and the Ph.D. degree in computer science and technology from BUPT in 2023.
He is currently an Associate Professor with the School of Cyberspace Science and Technology, Beijing Jiaotong University. His research interests include network security, moving target defense, and federated learning.
\end{IEEEbiography}

\begin{IEEEbiography}
[{\includegraphics[width=1in,height=1.25in,clip,keepaspectratio]{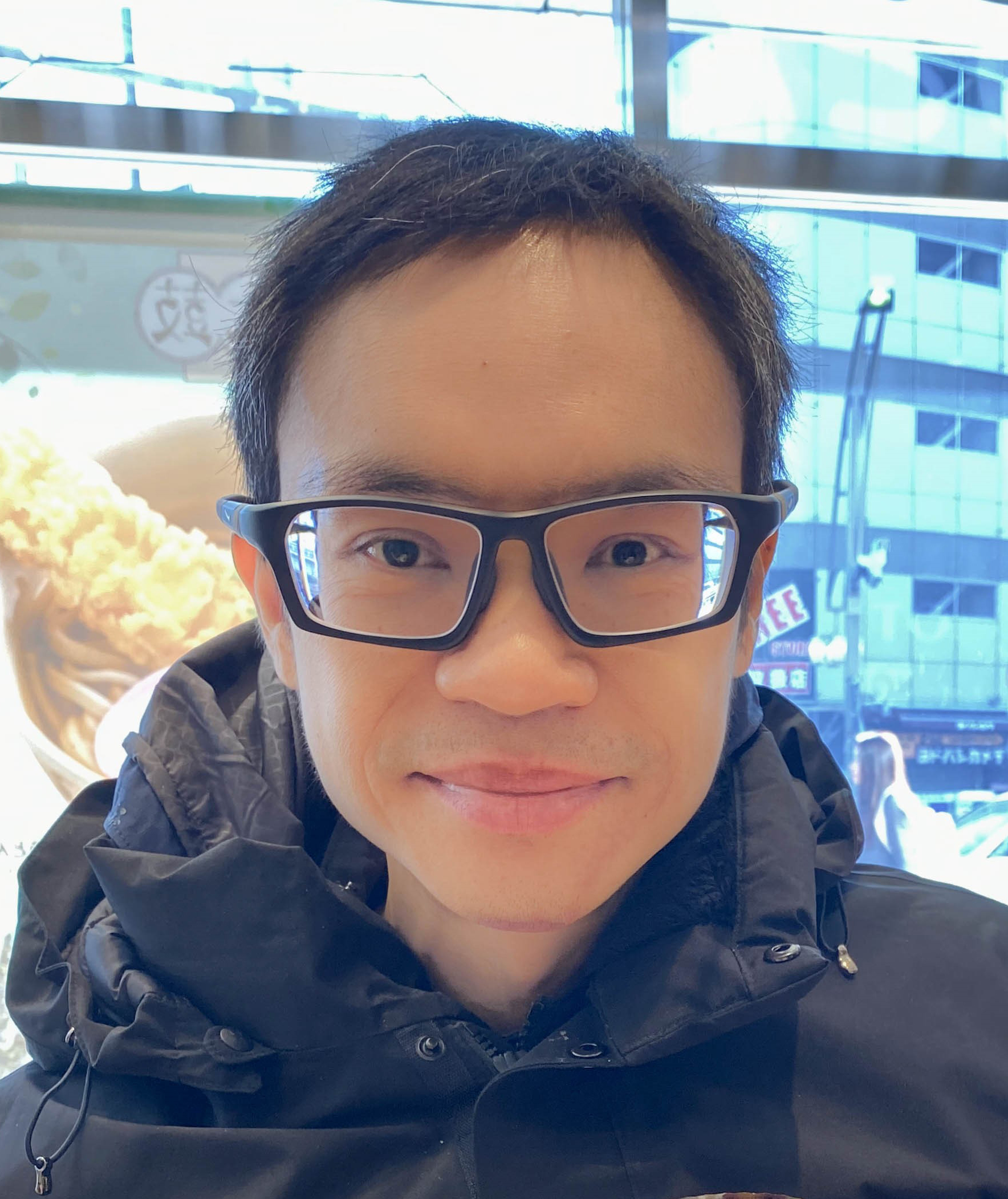}}] 
{Dusit Niyato} (M'09-SM'15-F'17) is a professor in the College of Computing and Data Science, at Nanyang Technological University, Singapore. He received B.Eng. from King Mongkuts Institute of Technology Ladkrabang (KMITL), Thailand and Ph.D. in Electrical and Computer Engineering from the University of Manitoba, Canada. His research interests are in the areas of sustainability, edge intelligence, decentralized machine learning, and incentive mechanism design.
\end{IEEEbiography}

\begin{IEEEbiography}
[{\includegraphics[width=1in,height=1.25in,clip,keepaspectratio]{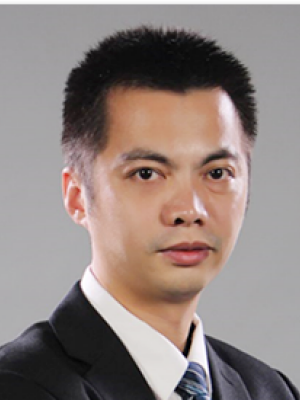}}] 
{Zibin Zheng} (Fellow, IEEE) received the Ph.D.
 degree from the Chinese University of Hong Kong, Hong Kong, in 2012. He is currently a full Professor with the School of Software Engineering,
Sun Yat-Sen University, Guangzhou, China. His
research interests include service computing and
cloud computing. Prof. Zheng was the recipient of the Outstanding Ph.D. Dissertation Award of the Chinese University of Hong Kong in 2012, ACM
SIGSOFT Distinguished Paper Award at ICSE in
2010, Best Student Paper Award at ICWS2010,
and IBM Ph.D. Fellowship Award in 2010. He was a PC Member of IEEECLOUD, ICWS, SCC, ICSOC, and SOSE.
\end{IEEEbiography}

\end{document}